\documentclass[runningheads]{llncs}

\usepackage{eccv}

\usepackage{eccvabbrv}

\usepackage{graphicx}
\usepackage{booktabs}
\usepackage{bbm}
\usepackage{tabularx}
\usepackage{makecell}
\newcommand{\noi}[1]{\noindent\textbf{#1}}

\makeatletter
\renewcommand\paragraph{\@startsection{paragraph}{4}{\z@}%
  {0.75ex \@plus0.2ex}%
  {-1em}%
  {\normalfont\normalsize\bfseries\upshape}}
\makeatother

\usepackage[accsupp]{axessibility}  % Improves PDF readability for those with disabilities.

\usepackage{hyperref}

\usepackage{orcidlink}

\begin{document}

\title{COMPASS: Grounding Composition-Intent Guidance in Unified Multimodal Models} 

\titlerunning{COMPASS}

% Second Author\inst{2,3}\orcidlink{1111-2222-3333-4444} \and
% Third Author\inst{3}\orcidlink{2222--3333-4444-5555}}

\author{Ziqi Zhou\inst{1,2}\orcidlink{0000-0001-8209-5055} \and Weize Quan\inst{2,3}\orcidlink{0000-0003-0892-581X}\thanks{Corresponding authors: \emph{qweizework@gmail.com, jun.zhoujun@antgroup.com}} \and
Mining Tan\inst{2,3}\orcidlink{0009-0006-1504-6169} \and Zhihan Chen\inst{2,3}\orcidlink{0009-0001-9213-6372} \and Dandan Zheng\inst{4}\orcidlink{0009-0005-2151-3547} \and Jingdong Chen\inst{4}\orcidlink{0000-0002-1872-2592} \and Jun Zhou\inst{4}\orcidlink{0000-0001-6033-6102}$^\star$ \and Weiming Dong\inst{2,3}\orcidlink{0000-0001-6502-145X} \and Dong-Ming Yan\inst{2,3}\orcidlink{0000-0003-2209-2404}}

% \author{Ziqi Zhou\inst{1,2} \and Weize Quan\inst{2,3}\thanks{Corresponding author: \emph{qweizework@gmail.com}} \and
% Mining Tan\inst{2,3} \and Zhihan Chen\inst{2,3} \and Dandan Zheng\inst{4} \and Jingdong Chen\inst{4} \and Jun Zhou\inst{4} \and Weiming Dong\inst{2,3} \and Dong-Ming Yan\inst{2,3}}

\authorrunning{Z. Zhou et al.}

%State Key Laboratory of Multimodal Artificial Intelligence Systems (
\institute{University of Edinburgh, United Kingdom \and
MAIS, Institute of Automation, Chinese Academy of Sciences, China
% \url{http://www.springer.com/gp/computer-science/lncs} 
\and
School of Artificial Intelligence, University of Chinese Academy of Sciences, China
\and
Ant Group, China}

\maketitle
\vspace{-2.5em}

\begin{abstract}
Composition is a high-level visual intent that governs where subjects are placed and how a scene is organized, yet current unified multimodal models remain unreliable at fine-grained composition recognition and struggle to turn such intent into controllable generation.
We present COMPASS, the first unified multimodal framework that grounds composition-intent control in a single system spanning both composition perception and composition-guided generation, with a shared expert token $\tau_c$ as the central intent anchor.
On the perception side, COMPASS injects composition expertise into an MoE backbone in a minimally invasive manner and distills the inferred intent into $\tau_c$.
On the generation side, COMPASS reuses $\tau_c$ as a global conditioning signal that steers the denoising trajectory, effectively converting passive composition analysis into explicit layout control.
To support systematic instruction-following composition learning and evaluation at scale, we construct \textsc{Comp-11}, a large-scale dataset with an 11-class taxonomy and reasoning-augmented annotations.
Extensive experiments show that COMPASS substantially improves category-level composition understanding and delivers more composition-consistent, prompt-faithful generation than strong baselines. The code and dataset for this work will be released \hyperlink{https://github.com/Jia1018/COMPASS}{here}. 
\end{abstract}

% \newpage

%
\begin{figure}[t]
    \centering
    \includegraphics[width=\textwidth]{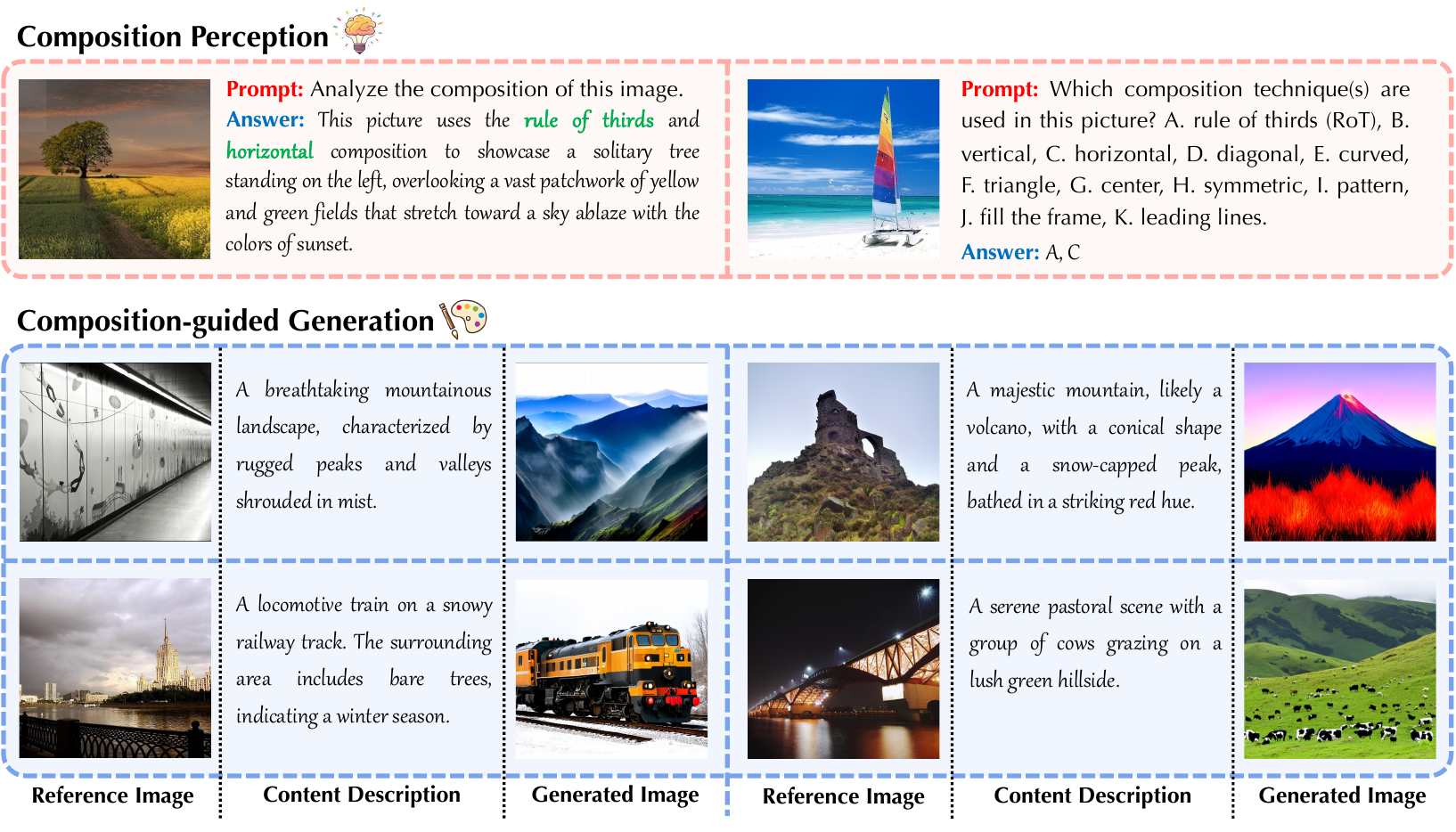}
    \caption{Performance of the proposed COMPASS: the unified system handles various composition understanding tasks, and further uses a reference image as layout intent to synthesize new content that remains faithful to the text prompt.}
    % \caption{Example performance of the proposed COMPASS system.}
    \label{fig:teaser}
\end{figure}
\vspace{-2.5em}

\section{Introduction}
\label{sec:intro}

Visual composition is the grammar of photography, providing the structural scaffold upon which aesthetic and emotional narratives are built. In the field of aesthetic intelligence, this necessitates two core capabilities working in tandem: expert-level perception, the ability to decode the intentional geometric organization of a scene, and layout-guided creation, the ability to synthesize new content that respects a specific compositional schema. Recently, Large Multimodal Models (LMMs) \cite{chameleon2024, deng2025emerging, gong2025minglite, emu3_2024, wu2025janus, xie2025showo, zhou2025transfusion} have trended towards a unified paradigm, aiming to integrate visual understanding and generation into a single, cohesive framework. However, while these models excel at general reasoning, they struggle to represent composition as an explicit, controllable structure. Existing aesthetic LMMs \cite{cao2025artimuse, huang2024aesexpert, zhou2024uniaa, cao2025unipercept} primarily emphasize passive assessment and critique, treating composition as one attribute among many rather than a structured object that can be parsed and acted upon. Meanwhile, the broader literature on photographic composition has proposed rule-based taxonomies and datasets \cite{lee2018photographic, zhang2021cadb, zhao2025picd}, which underscore that composition understanding is a distinct capability and remains underexplored for general LMMs.

In this work, we make the first attempt to unify expert composition perception and reference-guided generation in a cohesive framework. On the generation side, we target a challenging setting: given a reference image that conveys compositional intent and a text prompt for new semantic content, we seek to generate an image that inherits the reference layout style while changing the content. This pursuit is motivated by the inherent limitation of text-only prompting—language efficiently describes semantics but is a low-bandwidth channel for precise spatial geometry. In contrast, a reference photo provides a high-bandwidth spatial blueprint. Related controllable generation methods often rely on explicit low-level controls such as edges, depth, or bounding boxes \cite{Zhang_2023_ICCV, Li_2023_CVPR, zhang2023layoutdiffusion_iccv}, which provide strong spatial constraints but do not directly encode high-level composition rules. Closer in spirit, image-prompt editing and style-transfer methods \cite{ye2023ipadapter, Chung_2024_CVPR, wang2024instantstyle} also condition on a reference image and aim to transfer a specific factor (e.g., style) to a new image; however, they often suffer from content leakage from the reference ones. In our setting, the target factor—composition—is even more entangled with image content, making layout transfer without semantic contamination substantially harder.

Crucially, reference-guided compositional generation presupposes expert perception; the model must infer the structural intent of a reference before it can reproduce it. However, achieving this dual intelligence in a unified model raises two domain-specific hurdles. 
On the perception side, the challenge is twofold. 
% First, there is a severe scarcity of systematic data: existing aesthetic datasets lack the fine-grained, instruction-following taxonomy required to train models on the myriad geometric relationships inherent in photography. 
First, there is a severe scarcity of systematic compositional data: existing aesthetic datasets for LMM \cite{huang2024aesbench, huang2024aesexpert, zhou2024uniaa, cao2025artimuse, cao2025unipercept} often treat composition as a coarse-grained attribute within a broad aesthetic analysis, lacking a comprehensive and consistent taxonomy to cover the diverse layout patterns used in professional photography.
Second, injecting such specialized expertise into a pre-trained LMM through full-parameter fine-tuning or task-agnostic adapters like LoRA \cite{hu2022lora} can easily bias or overwrite general reasoning unless the expertise is explicitly anchored and routed. On the generation side, the task faces an entanglement trap where models struggle to disentangle the high-level compositional intent from low-level content features. This is especially difficult because paired data—images sharing the same layout but different semantics—rarely exists at scale, making conventional supervised disentanglement infeasible.

To address these challenges, we introduce COMPASS, a framework that empowers unified LMMs with specialized compositional intelligence through an expert-anchor paradigm. To overcome the scarcity of systematic data, we first build Comp-11, a comprehensive composition-focused instruction dataset. Inspired by prior composition datasets \cite{lee2018photographic, zhang2021cadb} and recent benchmarks highlighting LMMs’ compositional limitations \cite{zhao2025picd}, \textsc{Comp-11} is designed specifically for instruction-following and actionable composition understanding.
In terms of architecture, we propose a Composition-specific Mixture-of-Experts (C-MoE) strategy to achieve expert-level composition perception. Leveraging an existing MoE-based backbone \cite{gong2025minglite}, we instantiate a dedicated set of new expert Multilayer Perceptrons (MLPs) and optimize only these parameters during training. This approach infuses the model with compositional expertise while preserving its foundational multimodal competence. Complementing this architecture, we introduce a single learnable expert token $\tau_c$ as a prefix to the model's response. Supervised by a hard composition classification objective, $\tau_c$ serves as an explicit intent anchor that activates specialized knowledge and surfaces compositional decisions.

To break entanglement in generation without requiring paired data, we pioneer a self-supervised structural bottlenecking strategy. We apply explicit physical decoupling via pixelization and grayscaling to reference images, creating a structural invariant that suppresses semantic appearance while preserving global layout cues. Beyond the input-side bottleneck, we introduce model-side mechanisms to suppress reference leakage, including a customized attention mask that prevents learnable queries from attending to perturbed image tokens and a cross-attention refinement to ensure text faithfulness. Reusing the expert token $\tau_c$ as a global conditioning signal further bridges the gap between passive composition analysis and active, layout-controllable generation within a single unified system.

Our main contributions are summarized as follows: (1) We present the first systematic study of unified LMMs for compositional intelligence. (2) We build \textsc{Comp-11}, the first large-scale comprehensive composition-focused instruction dataset with an 11-class taxonomy to enable actionable understanding and reasoning. (3) We propose COMPASS, featuring a learnable prefix token $\tau_c$ as a shared intent anchor for both understanding and generation tasks, and a self-supervised structural bottlenecking framework that enables reference-guided generation without the need for paired training images.

\begin{figure}[t]
    \centering
    \includegraphics[width=\linewidth]{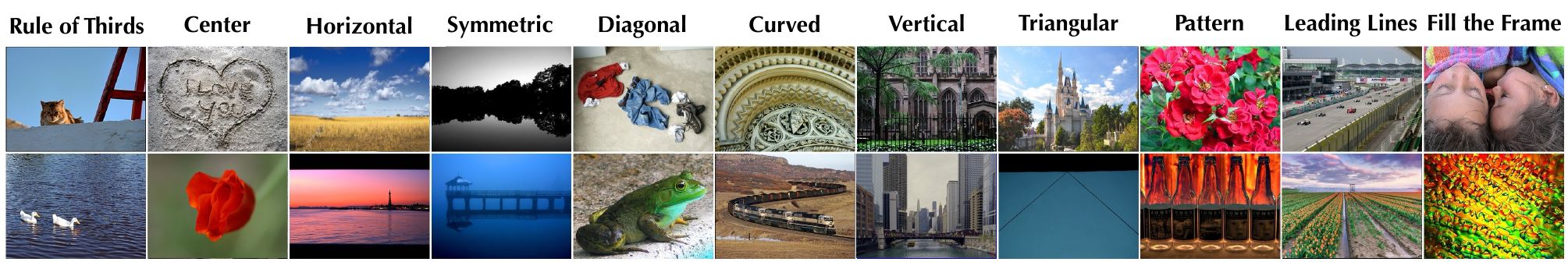}
    \caption{Visual examples of our \textsc{Comp-11} taxonomy.}
    \label{fig:taxonomy}
\end{figure}
    
\section{Related Work}
\paragraph{Composition understanding and optimization.}
Photographic composition has primarily been studied from two complementary perspectives: composition understanding and composition optimization. Together, these lines of research provide essential foundations for controllable composition generation.

The first line of research focuses on composition understanding, aiming to model and quantify structural layouts. Early approaches typically evaluated composition quality through regression-based models. By utilizing datasets with dedicated composition-aware annotations, such as CADB \cite{zhang2021cadb}, these methods isolated spatial arrangement patterns rather than predicting holistic aesthetic scores. Progressing toward a more structured formulation, subsequent studies treated composition as a rule-based classification task. For instance, KU-PCP \cite{lee2018photographic} categorizes images based on predefined compositional rules while explicitly localizing key geometric elements. Together, these works established quantitative and interpretable paradigms for composition recognition. 

Building upon composition evaluation, composition optimization focuses on active enhancement, primarily through image cropping. Existing methods are largely supported by two types of datasets: densely annotated datasets with multiple candidate crops (e.g., SACD \cite{yang2023focusing}, UGCrop5K \cite{su2024spatial}) and sparsely annotated datasets providing a single optimal crop (e.g., FCDB \cite{chen-wacv2017}). Recent advances incorporate generative modeling to improve robustness, as exemplified by GenCrop \cite{hong2024learning}, which leverages diffusion-based augmentation. Multimodal large models have also been introduced into composition tasks; for instance, PhotoFramer \cite{you2025photoframer} employs multimodal instruction tuning to generate composition guidance alongside exemplar images.

Despite these advances, existing optimization methods—particularly cropping—are inherently restricted to being post-processing steps for existing images. In contrast, our work transcends this limitation by providing real-time guidance, directly generating photographs with ideal compositions.

\paragraph{LMM aesthetic perception.}
More recently, multi-dimensional aesthetic evaluation has increasingly incorporated Large Multimodal Models (LMMs) to enhance semantic reasoning and interpretability \cite{achlioptas2021artemis}. Representative works—such as AesExpert \cite{huang2024aesexpert}, UNIAA \cite{zhou2024uniaa}, AesBench \cite{huang2024aesbench}, and UniPercept \cite{cao2025unipercept}—simultaneously generate aesthetic scores and natural language explanations. This approach facilitates unified reasoning across various aesthetic attributes, including color, lighting, subject emphasis, and composition.

However, within this paradigm, composition is primarily treated as a descriptive attribute embedded within a holistic aesthetic assessment. Even when compositional aspects are explicitly evaluated \cite{cao2025artimuse,huang2024aesexpert}, they are subsumed into the broader semantic reasoning process rather than being modeled as explicit, controllable composition categories. Consequently, existing LMM-based aesthetic perception frameworks remain predominantly evaluative, falling short of directly supporting composition-conditioned generation.

\paragraph{Unified multimodal models.}
Unified multimodal models seeks to integrate perception and generation within a shared architecture. Existing approaches can be broadly grouped into three paradigms. Modular coordination frameworks, such as NExT-GPT \cite{wu2024next}, employ a large language model to orchestrate external modality-specific generators. While flexible, these designs rely on loosely coupled components with limited cross-modal fusion. Token-level fusion models, including SEED-X \cite{ge2024seedx} and Chameleon \cite{chameleon2024}, move toward tighter integration by jointly modeling visual and textual tokens within autoregressive transformers, enabling unified understanding and generation. However, purely autoregressive modeling often struggles with high-fidelity image synthesis. To address this limitation, hybrid unified frameworks incorporate diffusion mechanisms into the modeling pipeline. Transfusion \cite{zhou2025transfusion} integrates diffusion into unified training to balance semantic controllability and synthesis quality. Subsequent models, such as Show-o \cite{xie2025showo}, Emu3 \cite{emu3_2024}, and the Janus series \cite{chen2025janus,wu2025janus,ma2024janusflow}, further extend unified architectures to support multi-task perception and high-quality generation. More recent efforts emphasize scalability and system-level refinement. VILA-U \cite{wu2024vila} improves multimodal alignment efficiency for large-scale training, while BAGEL \cite{deng2025emerging} and Ming-Lite-Uni \cite{gong2025minglite} explore scalable and lightweight unified modeling strategies.

Despite architectural unification, existing models primarily optimize modality alignment, generation quality, and training scalability \cite{transformervisual}. Fine-grained structural factors—such as photographic composition—are rarely treated as explicit, controllable modeling targets. Consequently, while unified multimodal architectures provide a strong backbone for integrated perception and generation, they lack dedicated mechanisms for structure-aware and composition-conditioned synthesis.

\begin{figure}[t]
    \centering
    \includegraphics[width=\linewidth]{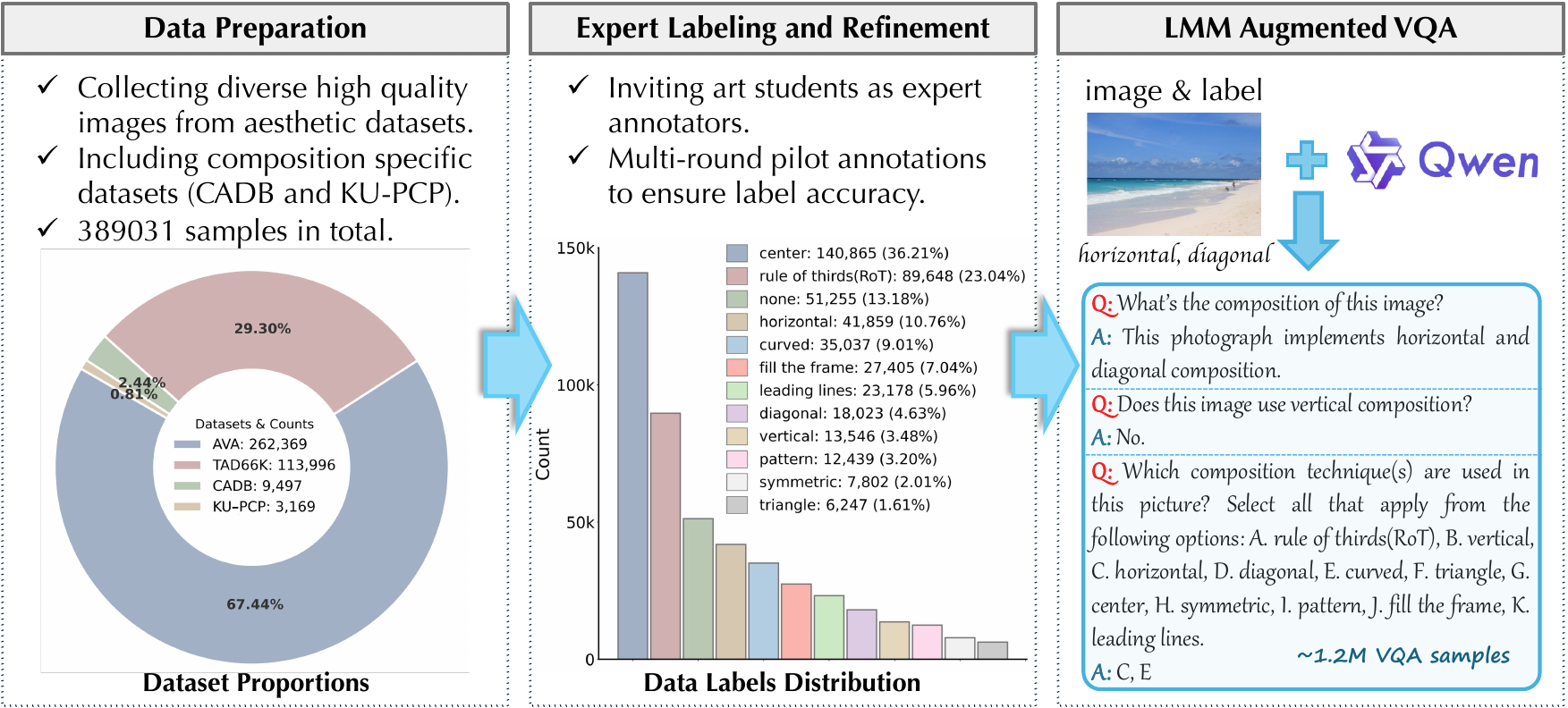}
    \caption{Data construction pipeline of \textsc{Comp-11}. }
    \label{fig:data_pipeline}
\end{figure}

\section{Dataset}
\label{sec:data}
We introduce \textsc{Comp-11}, a large-scale dataset for composition perception and generation in unified LMMs.
\textsc{Comp-11} is built upon an 11-class composition taxonomy and further augmented with instruction-following and reasoning-oriented multimodal annotations.

\paragraph{Taxonomy of Composition.}
We consolidate composition concepts from existing composition classification datasets KU-PCP~\cite{lee2018photographic} and CADB~\cite{zhang2021cadb} into a unified taxonomy with $C{=}11$ commonly used categories, which we adopt as the label set of our dataset.
Specifically, the taxonomy includes Rule of Thirds (RoT), Center, Horizontal, Symmetric, Diagonal, Curved, Vertical, Triangle, Pattern, Leading Lines, and Fill the Frame.
We provide qualitative visualization examples for all 11 categories in Figure~\ref{fig:taxonomy}.
Precise operational definitions and annotation guidelines are deferred to the supplementary material.

\paragraph{Data Construction Pipeline}
Building upon our 11-class taxonomy, we develop a rigorous pipeline comprising three stages: data preparation, expert labeling and refinement, and LMM augmentation (Figure~\ref{fig:data_pipeline}).

\emph{Data preparation.}
We aggregate 389,031 images from four public aesthetic benchmarks: AVA~\cite{murray2012ava}, TAD66K~\cite{he2022rethinking}, CADB~\cite{zhang2021cadb}, and KU-PCP~\cite{lee2018photographic}.
Among them, CADB and KU-PCP provide composition labels that can be directly reused, but their scale is insufficient for training a unified LMM.
To build a large and systematic training set, we further incorporate high-quality images from AVA and TAD66K as unlabeled aesthetic data and annotate them with our 11-category taxonomy in the next stage. The proportion of images contributed by each source can be seen from the left side of Figure~\ref{fig:data_pipeline}.

\emph{Expert labeling and refinement.} 
To ensure expert-level annotation quality, we recruited students from photography and fine arts programs and refined labels in two rounds: (i) verifying and correcting inherited annotations from source datasets, and (ii) exhaustively annotating newly collected data under our 11-class taxonomy.
To mitigate individual bias, each image is independently reviewed by multiple annotators and disagreements are resolved by consensus voting, yielding a unified and reliable label space.
The middle part of Figure~\ref{fig:data_pipeline} further indicates a long-tailed, highly imbalanced label distribution that naturally arises in real-world composition patterns.
Notably, we also allow a none-label assignment when an image is visually cluttered or its composition is indiscernible even for experts, avoiding forced noisy labels and keeping supervision reliable.

\emph{LMM augmentation.} Using the expert-verified labels as gold-standard prompts, we leverage a strong LMM~\cite{qwen2.5_vl_2025} to generate reasoning-enhanced metadata, including (i) question answering on compositional rationale, (ii) multiple-choice layout discrimination, and (iii) judgment tasks.
In total, this augmentation yields on the order of $1.2$M VQA-style pairs, substantially scaling up composition supervision for unified LMM training.

\begin{figure*}[t]
    \centering
    \includegraphics[width=\textwidth]{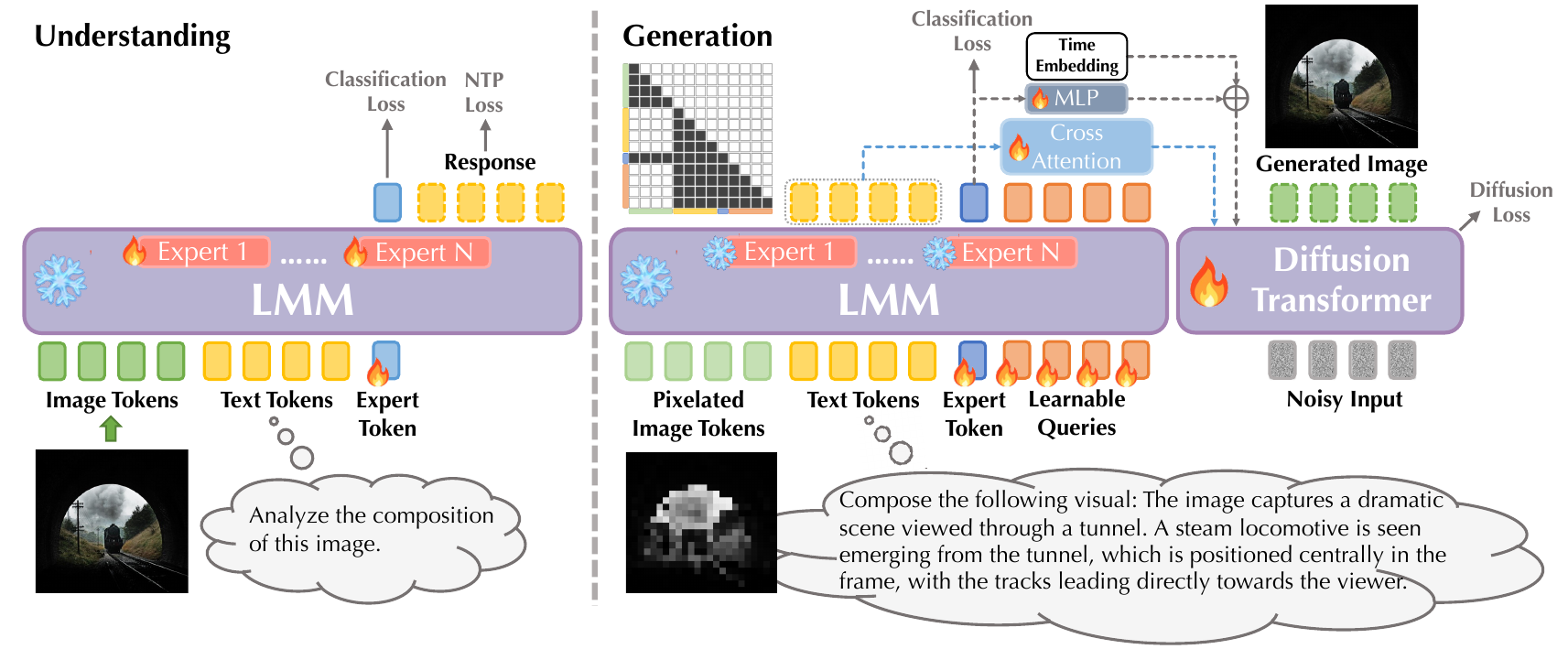}
    \caption{Overview of COMPASS. The model first performs composition perception with N new experts (C-MoE) and an expert token $\tau_c$ to summarize layout cues. For composition-guided generation, a structurally bottlenecked (pixelated) reference is used to suppress appearance leakage, while a customized attention mask (demonstrated under the bold ``Generation'', where the dark squares represent ``allow to attend'', while the white ones indicate the opposite) and a cross-attention refinement improve controllability and text faithfulness. The expert token representation is further injected into the diffusion denoiser as a global conditioning signal to steer the generation toward the intended composition.}
    \label{fig:overview}
\end{figure*}

\section{Methodology}
\label{sec:method}
With the dataset and supervision described in Section~\ref{sec:data}, we study compositional intelligence in unified large multimodal models through two tightly coupled tasks.
In composition perception, given an input image $I$ and an instruction, the model produces an instruction-following response and predicts a multi-label composition vector $y\in\{0,1\}^{C}$ (with $C=11$ in our setting).
In reference-guided compositional generation, given a reference image $I^{\mathrm{ref}}$ that provides layout intent and a text prompt $t$ that specifies new semantic content, we aim to sample an output image $I^{\mathrm{gen}}$ that preserves the reference composition while remaining faithful to the prompt:
\begin{equation}
I^{\mathrm{gen}} \sim \mathcal{G}\!\left(I^{\mathrm{ref}}, t\right),
\quad \text{s.t. } \mathrm{Comp}(I^{\mathrm{gen}})\approx \mathrm{Comp}(I^{\mathrm{ref}}),~
\mathrm{Sem}(I^{\mathrm{gen}})\approx t.
\end{equation}
To support these objectives within a single system, we introduce a minimally invasive perception expertization mechanism (C-MoE with a learnable expert token) and a self-supervised generation design that avoids paired same-layout-different-content supervision.
We describe the perception branch, the generation branch, and our training strategy as follows.

\subsection{Composition Perception}\label{sec:cmoe}
Compositional cues are subtle, geometry-centric, and poorly aligned with generic pretraining objectives, making naïve fine-tuning prone to either underfitting or overwriting general multimodal competence.
To this end, we design a perception branch with two complementary components: C-MoE, which injects composition expertise into the MoE backbone in a minimally invasive manner, and a learnable expert token $\tau_c$, which serves as an explicit intent anchor whose representation can be directly supervised for composition prediction.

\paragraph{Composition expertization via C-MoE.}
Our original unified LMM backbone \cite{gong2025minglite} adopts MoE-based feed-forward networks (FFNs) in a subset of transformer blocks.
Given latent token features $h\in\mathbb{R}^{d}$, the router first produces logits $z=g(h)\in\mathbb{R}^{N}$ over $N$ experts and selects the index set $\mathcal{K}=\mathrm{TopK}(z)$ of the $K$ largest entries in $z$ (Top-$K$ expert selection).
The combining weights are then computed by a softmax restricted to $\mathcal{K}$:
\begin{equation}
\alpha_k(h)=
\begin{cases}
\displaystyle \frac{\exp(z_k)}{\sum_{j\in\mathcal{K}}\exp(z_j)}, & k\in\mathcal{K},\\[6pt]
0, & k\notin\mathcal{K},
\end{cases}
\end{equation}
and the MoE-FFN output is a weighted sum of the selected experts:
\begin{equation}
\mathrm{MoE}(h)=\sum_{k=1}^{N} \alpha_k(h)\,E_k(h)=\sum_{k\in\mathcal{K}} \alpha_k(h)\,E_k(h),
\end{equation}
where $E_k(\cdot)$ denotes the $k$-th expert MLP.

To inject composition expertise without overwriting the foundation capabilities, we propose C-MoE (Composition-specific Mixture-of-Experts), which allocates a new MoE-FFN branch in each MoE block.
Concretely, for every sparse FFN layer we instantiate an additional MoE module,
named $\mathrm{MoE}^{\mathrm{comp}}(\cdot)$, with its own router and expert parameters, in parallel to the original backbone MoE,
noted as $\mathrm{MoE}^{\mathrm{base}}(\cdot)$.

At runtime, we switch between the two branches using a task flag $s\in\{0,1\}$:
\begin{equation}
\mathrm{FFN}(h) = (1-s)\,\mathrm{MoE}^{\mathrm{base}}(h) \;+\; s\,\mathrm{MoE}^{\mathrm{comp}}(h),
\end{equation}
where $s=1$ activates the composition expert branch.
Importantly, $\mathrm{MoE}^{\mathrm{comp}}$ may adopt a different routing policy (router type) from the original backbone,
enabling task-specific routing while keeping the backbone intact.

During training, we freeze the original backbone parameters
and update only the newly introduced parameters in $\mathrm{MoE}^{\mathrm{comp}}$ (its router and expert MLPs) to add compositional knowledge with minimal interference to the model's general multimodal abilities.

\paragraph{Learnable expert token as an intent anchor.}
To turn composition from an implicit aesthetic attribute into an explicit and actionable decision in unified LMM, we introduce a single learnable expert token $\tau_c$ as an intent anchor for composition understanding.
Concretely, given a dialogue context $\mathbf{x}$ formatted by a chat template, the assistant turn is constructed as
$\mathbf{x}\,\Vert\,\textit{prefix}_{\textit{ans}}\,\Vert\,\tau_c\,\Vert\,\mathbf{s}$,
where $\textit{prefix}_{\textit{ans}}$ denotes the fixed answer prefix produced by the template (e.g., role/format tokens) and $\mathbf{s}=s_{1:T}$ is the response with $T$ tokens.
We insert $\tau_c$ right after $\textit{prefix}_{\textit{ans}}$ and before the first response token, so that it functions as the answer-leading token and concentrates compositional cues into a dedicated representation for downstream supervision.

We supervise $\tau_c$ with a composition classification objective.
Let $H\in\mathbb{R}^{L\times d}$ be the final hidden states of the unified LMM for an input sequence of length $L$.
We extract the hidden representation at the expert-token position, denoted as $h_c\in\mathbb{R}^{d}$, and apply a lightweight projection head $\phi(\cdot)$ and a classifier $W$ to obtain logits $o\in\mathbb{R}^{C}$ over $C$ composition classes:
\begin{equation}
o = W\,\phi(h_c).
\end{equation}
Since an image may exhibit multiple composition techniques, we formulate composition prediction as multi-label classification and optimize a weighted binary cross-entropy loss:
\begin{equation}
\mathcal{L}_{\mathrm{cls}}
=
\sum_{i=1}^{C}
\left(
w_i^{+}\, y_i \log \sigma(o_i)
+
w_i^{-}\, (1-y_i)\log (1-\sigma(o_i))
\right),
\label{eq:weighted-bce}
\end{equation}
where $y\in\{0,1\}^{C}$ is the multi-hot label vector, $\sigma(\cdot)$ is the sigmoid function, and $(w_i^{+}, w_i^{-})$ are per-class weights to counter class imbalance (computed from the label distribution; see Section~\ref{sec:data}).
In practice, we optionally apply a pooling operator when multiple $\tau_c$ instances appear (e.g., in multi-round dialogues), but Eq.~\eqref{eq:weighted-bce} remains unchanged.

\subsection{Composition-guided Generation}
\paragraph{Physical decoupling via structural bottlenecking.}
In reference-guided compositional generation, the reference image only provides layout intent but not content.
Ideally, one would learn this factorized transfer from paired supervision---images sharing the same composition but different semantics---yet such paired data is scarce in practice.
Motivated by common self-supervised learning principles, we instead create a training signal by transforming each sample into a structural proxy that approximates ``same layout, different appearance'' without explicit pairs. 
% Since composition is highly entangled with semantics, purely implicit disentanglement is difficult and models tend to copy unintended content from the reference.

\begin{figure}[t]
    \centering
    \includegraphics[width=\linewidth]{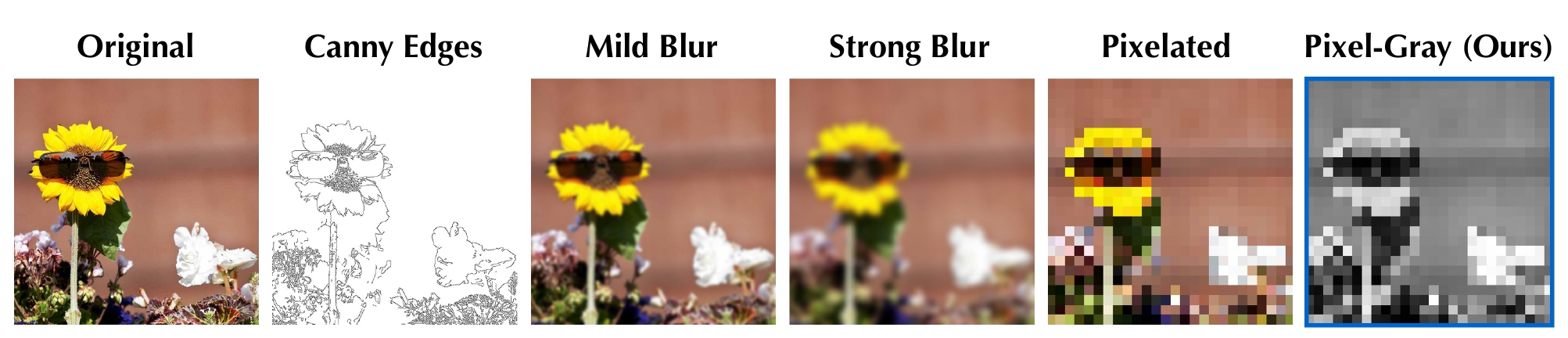}
    \caption{Comparison of structural bottlenecks for reference-guided generation. Pixelated grayscale references suppress semantic identifiability by disrupting contour-level cues while preserving coarse spatial layout, leading to better layout transfer with reduced content leakage compared to edge- or blur-based alternatives.}
    \label{fig:pix}
\end{figure}

Concretely, we introduce a structural bottleneck before the model: we transform the reference image $I^{\mathrm{ref}}$ into a grayscale, pixelated image $\tilde{I}$ and condition generation on $\tilde{I}$ instead of the raw reference.
Grayscaling suppresses color as a strong content cue, while pixelization aggressively removes recognizable details yet preserves coarse spatial massing and dominant layout lines.
% Alternative structuralizations exhibit an unfavorable trade-off: edge maps discard region-level composition cues and can over-emphasize local contours, whereas strong Gaussian blur can remove semantics but also washes out structural boundaries critical for composition transfer.
% Empirically (Figure~\ref{fig:xxx}), pixelized grayscale references yield the lowest content identifiability while maintaining sufficient layout signals for controllable generation.
Alternative structuralizations do not reliably eliminate semantics.
Edge-based representations preserve fine contours that directly encode object identity, while Gaussian blurring mainly smooths textures but often retains coherent silhouettes, allowing viewers (and models) to infer plausible semantics from shape cues even when details are unclear.
Pixelization offers a better compromise: it disrupts contour-level recognizability and thus weakens semantic inference, yet largely preserves coarse spatial layout and region massing.
Accordingly, we adopt grayscale pixelization as the structural bottleneck for reference-guided generation (Figure~\ref{fig:pix}).

\paragraph{Customized attention mask for leakage suppression.}
On top of the above input-side bottleneck, we further introduce model-side mechanisms to prevent reference content leakage and to
ensure text faithfulness during generation.
Following existing unified model architectures \cite{tong2024metamorph, wu2025janus, gong2025minglite}, when processing $\tilde{I}$, our unified LMM forms an autoregressive token sequence that contains (i) perturbed image tokens $\mathbf{v}=v_{1:M}$ extracted from $\tilde{I}$, (ii) text tokens $\mathbf{t}=t_{1:L}$, and (iii) a set of learnable queries $\mathbf{q}=q_{1:Q}$ used as the downstream generation interface.
Although $\tilde{I}$ already removes most semantic details, allowing text tokens or learnable queries to attend to
$\mathbf{v}$ can still create shortcut copying.
We therefore use a customized causal mask $A\in\{0,1\}^{N\times N}$ ($N$ is the model length) such that
\begin{equation}
A_{ij}=\mathbbm{1}[j\le i]\cdot\mathbbm{1}\big[(i\notin \mathcal{T}\cup\mathcal{Q}\setminus\{\tau_c\})\ \vee\ (j\notin \mathcal{I})\big],
\label{eq:attention_mask}
\end{equation}
where $\mathcal{I}$ indexes the reference-image token span, $\mathcal{Q}$ indexes the learnable-query span, and $\mathcal{T}$ indexes all text-token positions after the reference image.
Thus, standard causality ($j\le i$) is preserved, while any position $i\in(\mathcal{T}\cup\mathcal{Q})\setminus\{\tau_c\}$ is forbidden to attend to reference-image tokens $j\in\mathcal{I}$, leaving $\tau_c$ as the only post-image token that may read $\mathbf{v}$ for composition-related summarization.
We visualize the allowed and blocked regions of $A$ in Figure~\ref{fig:overview} (under the bold ``Generation'') for more intuitive demonstration.

\paragraph{Cross-attention refinement for text faithfulness.}
Reference-guided generation is prone to content leakage unless the model is simultaneously discouraged from copying the reference and encouraged to follow the prompt \cite{xu2025stylessp}. Our customized attention mask implements the former by preventing post-image tokens and learnable queries from attending to reference-image tokens. To complement this, we add a lightweight cross-attention refinement module that directly ties the learnable queries to the prompt representation, strengthening text faithfulness in the diffusion conditioning stream. Concretely, after the LMM forward pass, let $H\in\mathbb{R}^{N\times d}$ be the resulting hidden states. We take the subsequences for learnable queries and prompt tokens, denoted as $H_{\text{queries}}=\text{LMM}(\textbf{q})\in\mathbb{R}^{Q\times d}$ and $H_{\text{prompt}}=\text{LMM}(\textbf{t})\in\mathbb{R}^{L\times d}$, and apply
\begin{equation}
\hat{\textbf{q}}=\mathrm{CrossAttn}(H_{\text{queries}},\,H_{\text{prompt}}),
\label{eq:cross_attention}
\end{equation}
where $H_{\text{queries}}$ acts as queries and $H_{\text{prompt}}$ provides keys/values. This process anchors the learnable queries to the prompt representation and thereby improving text content faithfulness.

\paragraph{Expert token as a global conditioning signal.}
Finally, we reuse the expert token $\tau_c$ as a layout capsule that bridges composition perception and controllable generation.
In our generation setting, the reference-image pathway is deliberately weakened, so the model cannot reliably exploit residual appearance cues from reference tokens; instead, transferable layout information is encouraged to be summarized into the representation at $\tau_c$.
Formally, let $h_c\in\mathbb{R}^{d}$ be the final hidden state at $\tau_c$, and let $c=\psi(h_c)$ be a projection into the diffusion conditioning space.
We inject $c$ as a global bias via timestep modulation, $\tilde{e}_t = e_t + c$, where $e_t$ is the standard timestep embedding.
The modulated embedding $\tilde{e}_t$ is broadcast within the diffusion transformer and used to modulate denoising blocks in an AdaLN-style manner, supplying a scene-level control signal that steers the denoising trajectory toward the intended composition.
Importantly, we keep the composition classification supervision on $\tau_c$ enabled during generation training to handle the domain shift introduced by the perturbed reference images.
This encourages $\tau_c$ to remain a stable carrier of compositional cues and reduces the tendency to fall back to shortcut copying.

\subsection{Stage-wise Training Objective and Strategy}
\label{sec:training}

We train composition perception and reference-guided generation in two stages, rather than jointly, to stabilize optimization and to better match their heterogeneous input domains.

\paragraph{\textbf{Stage I: composition perception.}}
We optimize the unified LMM on the perception data with a standard instruction-following objective together with the composition supervision on the expert token:
\begin{equation}
\mathcal{L}_{\mathrm{perc}}
= \mathcal{L}_{\mathrm{ntp}} \;+\; \lambda_{\mathrm{cls}}\,\mathcal{L}_{\mathrm{cls}},
\end{equation}
where $\mathcal{L}_{\mathrm{ntp}}$ denotes the next-token prediction loss for autoregressive language models (details in the supplementary material) and $\mathcal{L}_{\mathrm{cls}}$ is the weighted multi-label classification loss defined in Eq.~\eqref{eq:weighted-bce}.
During this stage, we keep the backbone frozen and update only the composition-specific modules, including the C-MoE branch, the learnable expert token $\tau_c$ and its lightweight prediction head.

\paragraph{Stage II: reference-guided generation.}
We then train the generation pipeline under structural bottlenecking, where the model must follow the text prompt while transferring only layout cues from the perturbed reference.
The objective combines the diffusion training loss with the same $\mathcal{L}_{\mathrm{cls}}$ on $\tau_c$:
\begin{equation}
\mathcal{L}_{\mathrm{gen}}
= \mathcal{L}_{\mathrm{diff}} \;+\; \lambda'_{\mathrm{cls}}\,\mathcal{L}_{\mathrm{cls}}.
\end{equation}
Here $\mathcal{L}_{\mathrm{diff}}$ is the standard diffusion objective used by the denoiser (detailed in the supplementary material).
In this stage, we freeze the C-MoE composition experts learned in Stage~I to preserve the acquired perception capability and to maintain a single shared expertization module across tasks.
Meanwhile, we train all generation-specific parameters and re-train the expert token $\tau_c$ (and its lightweight head) to account for the domain shift of reference inputs. Although $\tau_c$ is trained separately across stages, it is lightweight and can be packaged with the unified system as task-specific parameters.

\begin{table*}[t]
\centering
\caption{Category-wise AP (\%) for composition understanding on the \textsc{Comp-11} perception test split.
$\dagger$ denotes understanding-only models, and $\ddagger$ denotes unified models.}
\label{tab:ap_comp11}
\resizebox{\textwidth}{!}{
\begin{tabular}{lcccccccccccc}
\toprule
Method
& RoT & Center & Hori. & Vert. & Diag. & Curved & Tri. & Sym. & Patt. & Lead. & Fill & mAP $\uparrow$ \\
\midrule
AesExpert$^{\dagger}$ \cite{huang2024aesexpert} & 42.3 & 50.6 & 53.0 & 50.2 & 47.8 & 50.4 & 63.9 & 56.3 & 54.4 & 58.1 & 50.1 & 49.5 \\
Qwen3-VL$^{\dagger}$ \cite{qwen3vl_2025} & \underline{56.7} & \underline{70.8} & 78.7 & \underline{76.9} & 54.1 & 54.6 & 87.2 & \underline{86.0} & 61.8 & 72.5 & \underline{66.2} & \underline{66.4} \\
InternVL3$^{\dagger}$ \cite{zhu2025internvl3} & 50.6 & 65.7 & 57.5 & 55.9 & 55.8 & \underline{64.3} & 82.8 & 78.3 & 65.5 & 62.3 & 52.6 & 60.0 \\
Janus-Pro$^{\ddagger}$ \cite{chen2025janus} & 50.2 & 55.1 & 56.7 & 66.8 & 52.7 & 51.5 & 71.0 & 69.5 & 55.5 & 51.5 & 52.3 & 54.2 \\
BAGEL$^{\ddagger}$ \cite{deng2025emerging} & 51.1 & 66.4 & \underline{80.2} & 68.2 & 54.2 & 57.9 & \underline{87.4} & 83.2 & \underline{70.7} & \underline{74.0} & 52.8 & 63.2 \\
Ming-Lite-Uni$^{\ddagger}$ \cite{gong2025minglite} & 31.4 & 68.1 & 67.4 & 63.2 & \underline{63.9} & 57.3 & 84.2 & 82.0 & 69.8 & 65.8 & 62.2 & 58.2 \\
\midrule
COMPASS (ours)$^{\ddagger}$ & \textbf{90.1} & \textbf{86.2} & \textbf{96.1} & \textbf{97.5} & \textbf{96.2} & \textbf{92.9} & \textbf{98.4} & \textbf{98.1} & \textbf{95.3} & \textbf{96.2} & \textbf{87.2} & \textbf{90.6} \\
\bottomrule
\end{tabular}}
\end{table*}

\section{Experiments}
\label{sec:experiments}

\paragraph{Experimental setup.}
With the constructed \textsc{Comp-11} dataset, we evaluate COMPASS on both composition perception and composition-guided generation.
For perception, we hold out 20\% of the full dataset as a test split; the class distribution of the test set is provided in the supplementary material.
For generation, we further sample 10,000 evaluation groups from the perception test split with a uniform coverage over composition categories.
Each group consists of (i) a reference image that provides layout intent and (ii) a content reference image whose caption is used as the semantic prompt to the model.
Additional implementation details are included in the supplementary material.

\begin{table}[t]
\centering
\caption{Quantitative results on composition-guided generation.
$\dagger$ denotes generation-only models, and $\ddagger$ denotes unified models.}
\label{tab:gen_compare}
\begin{tabular}{lcccc}
\toprule
Method & FID $\downarrow$ & CLIPScore $\uparrow$ & Comp-Cons. $\uparrow$ & $\cos(h_c^{\mathrm{ref}}, h_c^{\mathrm{gen}})$ $\uparrow$ \\
\midrule
Step1X-Edit$^{\dagger}$ \cite{liu2025step1x-edit} & \textbf{7.5} & 0.72 & 0.60 & 0.51 \\
BAGEL$^{\ddagger}$ \cite{deng2025emerging} & 7.8 & 0.76 & 0.59 & 0.48 \\
Ming-Lite-Uni$^{\ddagger}$ \cite{gong2025minglite} & 9.6 & 0.64 & 0.53 & 0.44 \\
\midrule
COMPASS (ours)$^{\ddagger}$ & 8.2 & \textbf{0.83} & \textbf{0.74} & \textbf{0.62} \\
\bottomrule
\end{tabular}
\end{table}

\begin{table}[t]
\caption{Ablation study on composition-guided generation. $\star$ denotes ``same as the previous row'' with additional removals.
$^{\mathrm{P}}$ and $^{\mathrm{G}}$ indicate perception-stage and generation-stage ablations, respectively.}
\label{tab:supp_ablation}
\centering
\small
\setlength{\tabcolsep}{3.5pt}
\renewcommand{\arraystretch}{1.08}
\begin{tabularx}{\columnwidth}{@{}>{\raggedright\arraybackslash}Xccccc@{}}
\toprule
Variant &
mAP $\uparrow$ &
FID $\downarrow$ &
CLIPScore $\uparrow$ &
Comp-Cons. $\uparrow$ &
$\cos(h_c^{\mathrm{ref}}, h_c^{\mathrm{gen}})$ $\uparrow$ \\
\midrule
Full (COMPASS) & \textbf{90.6} & 8.2 & \textbf{0.83} & \textbf{0.74} & \textbf{0.62} \\
w/o C-MoE$^{\mathrm{P}}$ & 78.2 & -- & -- & -- & -- \\
w/o $\tau_c$$^{\mathrm{P}}$ & 71.1 & -- & -- & -- & -- \\
w/o cross attention$^{\mathrm{G}}$ & -- & \textbf{8.0} & 0.79 & 0.64 & 0.56 \\
$\star$ + w/o mask$^{\mathrm{G}}$ & -- & 8.6 & 0.72 & 0.59 & 0.48 \\
$\star$ + w/o $\tau_c$$^{\mathrm{G}}$ & -- & 8.3 & 0.75 & 0.53 & 0.42 \\
\bottomrule
\end{tabularx}
\end{table}

\subsection{Quantitative results}
\label{sec:quant_results}

\paragraph{Metrics.}
We report results on three aspects.
(i) \emph{Category-level composition understanding}: we evaluate each composition category as a binary decision and report average precision (AP), which directly reflects the model's discriminative capability under class imbalance.
(ii) \emph{Composition-guided generation}: we measure general quality by FID \cite{heusel2017gans} and CLIPScore \cite{hessel-etal-2021-clipscore}, and measure composition transfer by the agreement (Comp-Cons.) between predicted composition labels (by COMPASS) of the reference and generated images and the cosine similarity ($\cos(h_c^{\mathrm{ref}}, h_c^{\mathrm{gen}})$) between their expert-token hidden representations. (iii) \emph{Head-only classification} (in supplementary material): we compute multi-class accuracy using the lightweight head attached to the expert token. Detailed definitions can be found in the supplementary material.

\paragraph{Baselines.}
For perception, we compare against strong multimodal models, including an open-source aesthetic finetuned expert model: AesExpert \cite{huang2024aesexpert}, state-of-the-art VLMs: Qwen3-VL (8B) \cite{qwen3vl_2025} and InternVL3 (8B) \cite{zhu2025internvl3}, and unified models: Janus-Pro (7B) \cite{chen2025janus}, BAGEL (7B) \cite{deng2025emerging}, and our backbone Ming-Lite-Uni (8B) \cite{gong2025minglite}.
For generation, we compare against image editing/unified models that support reference-guided generation, including Step1X-Edit \cite{liu2025step1x-edit}, BAGEL \cite{deng2025emerging}, and Ming-Lite-Uni \cite{gong2025minglite}.
When a baseline does not support pixelated grayscale references, we feed the original reference image and prepend the prompt with
\emph{``Generate an image with the composition style of the reference image and the following content: ''} for a fair comparison.

\paragraph{Main results.}
Table~\ref{tab:ap_comp11} reports category-level AP for composition understanding. We observe that even strong contemporary LMMs still struggle to reliably identify fine-grained composition types, with performance often varying substantially across categories, highlighting that composition-centric reasoning remains underdeveloped in general-purpose multimodal models.
In contrast, COMPASS consistently improves recognition across all categories, demonstrating that explicit composition expertization and direct supervision on the expert token can effectively elicit composition knowledge.
Table~\ref{tab:gen_compare} summarizes quantitative generation results, where the substantial gains in composition-specific metrics and CLIPScore confirm that our framework effectively inherits the reference layout intent while successfully mitigating semantic leakage from the reference image.
More importantly, by grounding generation on explicit composition understanding, COMPASS enables composition-intent controllable generation---an expert capability that is largely absent from current unified models.
% Overall, COMPASS achieves strong gains on both perception and generation, indicating that the proposed expertization and leakage-suppression mechanisms jointly improve compositional controllability and text faithfulness.

\paragraph{Ablations.}
We ablate key components in both perception and generation. The results are presented in Table~\ref{tab:supp_ablation}. 
% On perception, we remove C-MoE or the expert token.
% On generation, we remove the cross-attention refinement, then additionally remove the customized attention mask, and finally remove the expert-token conditioning.
% Results are summarized in Table~\ref{tab:ablation}.
On perception, removing either C-MoE or the expert token causes a pronounced drop in mAP, suggesting that composition recognition benefits from both (i) explicit capacity allocation for composition features (C-MoE) and (ii) a dedicated, directly supervised intent carrier ($\tau_c$).
On generation, removing the cross-attention refinement and the customized attention mask mainly harms text faithfulness, reflected by a decrease in CLIPScore. Finally, disabling the expert-token conditioning degrades composition-transfer metrics, confirming that $\tau_c$ serves as the global layout capsule that stabilizes composition control throughout the denoising trajectory.
Due to space limitations, detailed numerical results are provided in the supplementary material.

\begin{figure}[t]
  \centering
  \includegraphics[width=\columnwidth]{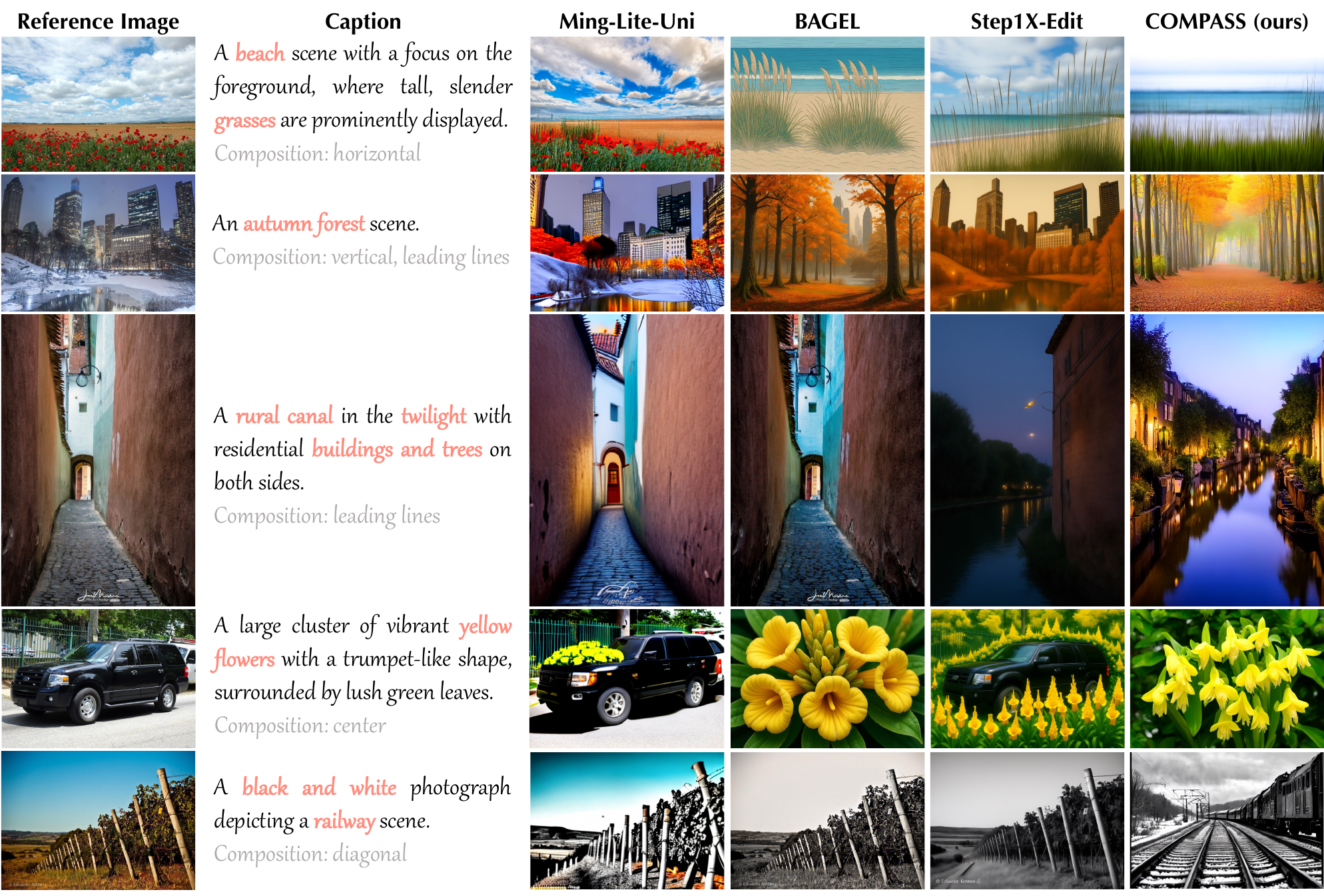}
  \caption{Qualitative comparison on composition-guided generation. Gray annotations under the captions are for intuitive illustration only and are not used as model inputs.}
  \label{fig:qual_gen}
\end{figure}

% Go to supp
% \begin{table}[t]
% \caption{Ablation study on composition-guided generation. $\star$ denotes ``same as the previous row'' with additional removals.
% $^{\mathrm{P}}$ and $^{\mathrm{G}}$ indicate perception-stage and generation-stage ablations, respectively.}
% \label{tab:ablation}
% \centering
% \small
% \setlength{\tabcolsep}{3.5pt}
% \renewcommand{\arraystretch}{1.08}
% \begin{tabularx}{\columnwidth}{@{}>{\raggedright\arraybackslash}Xccccc@{}}
% \toprule
% Variant &
% mAP $\uparrow$ &
% FID $\downarrow$ &
% CLIPScore $\uparrow$ &
% Comp-Cons. $\uparrow$ &
% $\cos(h_c^{\mathrm{ref}}, h_c^{\mathrm{gen}})$ $\uparrow$ \\
% \midrule
% Full (COMPASS) & 90.6 & -- & -- & -- & -- \\
% w/o C-MoE$^{\mathrm{P}}$ & -- & -- & -- & -- & -- \\
% w/o $\tau_c$$^{\mathrm{P}}$ & 71.1 & -- & -- & -- & -- \\
% w/o cross attention$^{\mathrm{G}}$ & -- & -- & -- & -- & -- \\
% $\star$ + w/o mask$^{\mathrm{G}}$ & -- & -- & -- & -- & -- \\
% $\star$ + w/o $\tau_c$$^{\mathrm{G}}$ & -- & -- & -- & -- & -- \\
% \bottomrule
% \end{tabularx}
% \end{table}

\subsection{Qualitative results}
Figure~\ref{fig:qual_gen} presents representative qualitative comparisons for composition-guided generation.
Rather than directly copying the reference layout, COMPASS first infers the compositional intent implied by the reference and then integrates this intent into the semantic content, yielding generations that are more harmonious and self-consistent while remaining faithful to the prompt semantics.
Additional qualitative examples for both composition perception and composition-guided generation, together with failure-case analysis, are provided in Figure~\ref{fig:teaser} and the supplementary material.

% % Go to supp
% % \begin{table*}[t]
% % \centering
% % \caption{Category distribution of the perception test split of \textsc{Comp-11}.}
% % \label{tab:test_dist}
% % \resizebox{\textwidth}{!}{
% % \begin{tabular}{lccccccccccc}
% % \toprule
% %  & RoT & Center & Hori. & Vert. & Diag. & Curv. & Tri. & Sym. & Patt. & Lead. & Fill \\
% % \midrule
% % \#Samples & -- & -- & -- & -- & -- & -- & -- & -- & -- & -- & --  \\
% % Ratio (\%) & -- & -- & -- & -- & -- & -- & -- & -- & -- & -- & --  \\
% % \bottomrule
% % \end{tabular}}
% % \end{table*}

% Go to supp
% \section{Discussion}
% 标签准确度以及美学评估的主观性

\section{Conclusion}
In this work, we present COMPASS, the first unified framework to bridge expert-level composition perception and reference-guided creation within a single LMM architecture. 
Supported by our systematic \textsc{Comp-11} dataset, COMPASS utilizes an expert-anchor paradigm and a self-supervised structural bottleneck to effectively decouple compositional intent from semantic content without requiring paired training data. 
This methodology transforms unified models from composition-blind observers into expert-level aesthetic creators, establishing a generalizable blueprint for domain-specific multimodal intelligence. 
% Future research will focus on extending this framework to encompass broader aesthetic attributes and investigating even more granular structural control mechanisms.

% \section*{Acknowledgements}
% Please insert your acknowledgments here.

\noi{Acknowledgment.}
This work was partially supported by the National Natural Science Foundation of China (62572469, 62572458) and the Ant Group.

\bibliographystyle{splncs04}
\bibliography{main}

\newpage
% =========================
% Supplementary Material
% =========================
\appendix
\section*{Appendix}

This appendix provides additional details that are omitted from the main paper due to space constraints, including: (i) taxonomy definitions and annotation guidelines, (ii) full objective formulations, (iii) evaluation metric definitions, (iv) ablation records, (v) additional qualitative results and failure cases, and (vi) ethics considerations.

% -------------------------------------------------
\section{\textsc{Comp-11} Taxonomy and Annotation Guidelines}
\label{sec:supp_taxonomy}

\subsection{Taxonomy Definitions}
\label{sec:supp_taxonomy_defs}

\begin{figure}[htbp]
  \centering
  \includegraphics[width=\columnwidth]{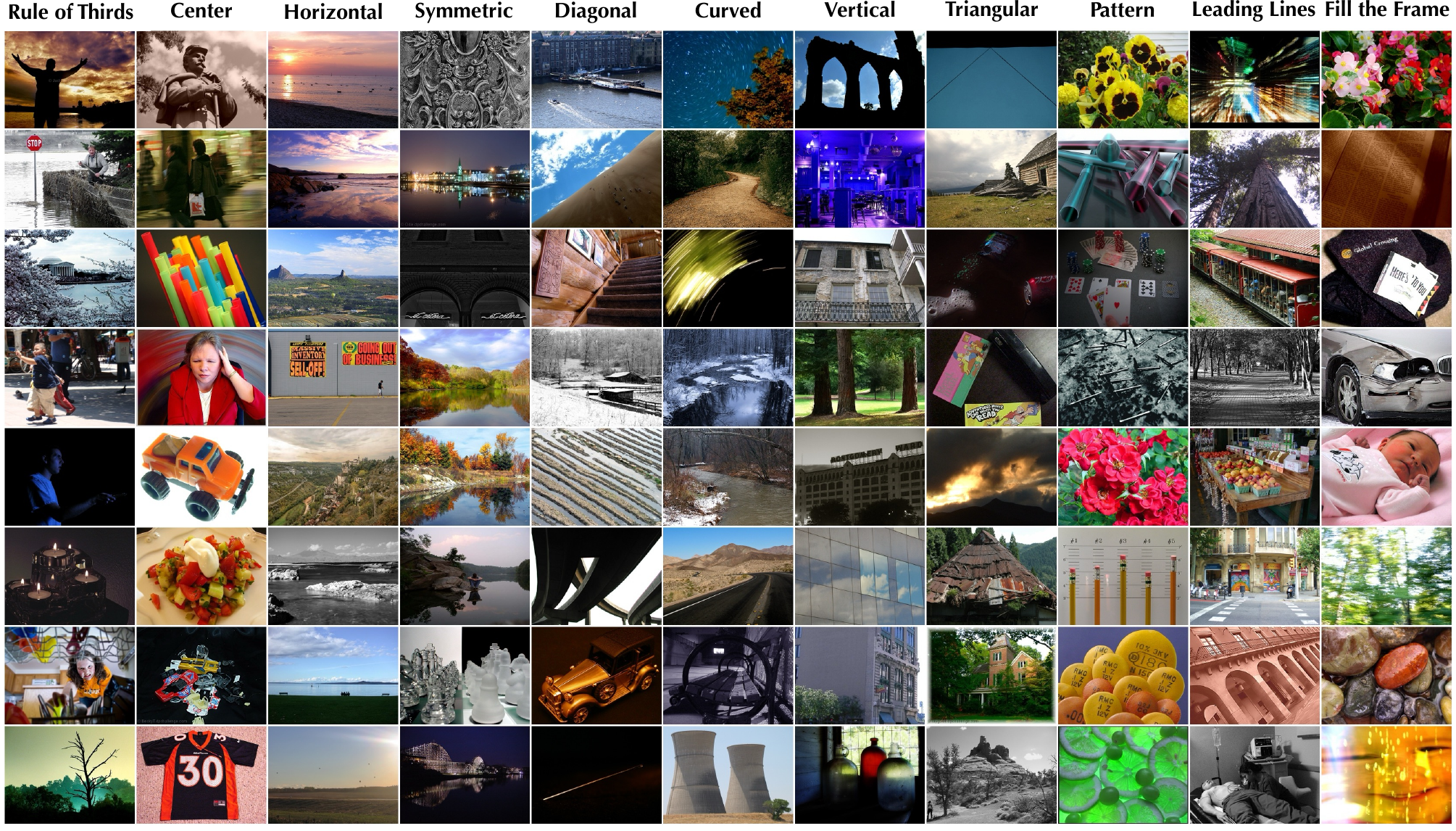}
  \caption{Typical examples for the 11 composition categories in \textsc{Comp-11}.}
  \label{fig:supp_taxonomy}
\end{figure}

We adopt an 11-category composition taxonomy: Rule of Thirds (RoT), Center, Horizontal, Vertical, Diagonal, Curved, Triangle, Symmetric, Pattern, Leading Lines, and Fill the Frame.
Each image may exhibit multiple composition techniques; hence labels are multi-hot vectors $y\in\{0,1\}^{11}$. We provide qualitative visualization examples for all categories in Figure~\ref{fig:supp_taxonomy}.

\paragraph{Rule of Thirds (RoT).}
The main subject(s) are placed near the intersections or lines of a $3\times3$ grid.

\paragraph{Center.}
The primary subject is centered, with salient mass concentrated around the image center.

\paragraph{Horizontal / Vertical.}
The global structure is dominated by horizontal (resp. vertical) alignments, e.g., horizons or vertical pillars.

\paragraph{Diagonal.}
The primary structure follows a dominant diagonal direction (top-left to bottom-right or top-right to bottom-left).

\paragraph{Curved.}
The layout exhibits salient curved flow (e.g., S-curves, arcs) that organizes subject placement or visual guidance.

\paragraph{Triangle.}
Key elements form a triangular arrangement (explicit or implicit), yielding a stable triangular composition.

\paragraph{Symmetric.}
The scene is organized with strong reflectional symmetry (horizontal or vertical), often centered around an axis.

\paragraph{Pattern.}
Repeated visual elements form a regular or quasi-regular spatial pattern.

\paragraph{Leading Lines.}
Prominent lines (e.g., roads, rails, edges) guide the viewer's gaze toward the main subject or vanishing point.

\paragraph{Fill the Frame.}
The subject occupies most of the frame (e.g., close-up composition) with minimal background distraction.

\subsection{Annotation Protocol}
\label{sec:supp_annotation_protocol}

\paragraph{Multi-label rules.}
Annotators assign all applicable categories when multiple composition techniques co-occur.

\paragraph{None-label (all-zero) rule.}
If an image is visually cluttered or its composition is indiscernible even for experts, we allow an all-zero assignment to avoid forcing noisy labels.

\paragraph{Conflict resolution.}
Each image is reviewed by multiple annotators. Disagreements are resolved by consensus voting under a cross-validation protocol.

\paragraph{Label distribution.}
As shown in Figure~\ref{fig:data_pipeline} in the main paper, the resulting taxonomy follows a long-tailed distribution in real-world data, reflecting the unequal prevalence of different composition patterns.
For evaluation, we construct the perception test split by holding out approximately 20\% of the dataset (with stratification over composition labels), and report its per-label prevalence in Table~\ref{tab:supp_test_dist}.

\begin{table*}[t]
\caption{Per-category test set sizes for binary composition recognition benchmarks (balanced positives/negatives by construction).}
\label{tab:supp_test_dist}
\centering
\setlength{\tabcolsep}{4pt}
\renewcommand{\arraystretch}{1.05}
\resizebox{\textwidth}{!}{
\begin{tabular}{lcccccccccccc}
\toprule
Category & RoT & Vert. & Hori. & Diag. & Curved & Tri. & Center & Sym. & Patt. & Fill & Lead. & Total \\
\midrule
\#Samples & 21{,}534 & 2{,}400 & 8{,}434 & 3{,}694 & 6{,}974 & 1{,}396 & 30{,}174 & 1{,}786 & 2{,}944 & 5{,}780 & 4{,}220 & 89{,}336 \\
\bottomrule
\end{tabular}}
\end{table*}

\subsection{Generation Evaluation Set Distribution}
\label{sec:supp_generation_eval_dist}
For the composition-guided generation task, to ensure an unbiased evaluation across different layout intents, the 10,000 evaluation image-prompt pairs are uniformly sampled. This results in approximately 909 samples per composition category, providing a balanced and comprehensive testbed for generation controllability.

% -------------------------------------------------
\section{Objectives and Optimization Details}
\label{sec:supp_objectives}

\subsection{Stage I: Next-Token Prediction Loss}
\label{sec:supp_ntp}

Let the tokenized training sequence be $u_{1:T}$ (including image/text special tokens according to the template).
The standard autoregressive next-token prediction (NTP) objective is
\begin{equation}
\mathcal{L}_{\mathrm{ntp}}
=
-\sum_{t=1}^{T}
\log p_\theta(u_t \mid u_{<t}).
\label{eq:supp_ntp}
\end{equation}
In practice, we apply standard label masking for non-target positions (e.g., prompt tokens).

\subsection{Stage II: Diffusion Training Objective}
\label{sec:supp_diff}

We use a standard diffusion/denoising objective.
Let $x$ be the target image and $z$ be its latent.
At timestep $t$, we form a noised latent $z_t$ and train a denoiser $\epsilon_\theta$ conditioned on the refined query embeddings and the expert-token conditioning:
\begin{equation}
\mathcal{L}_{\mathrm{diff}}
=
\mathbb{E}_{t,\epsilon}
\left[
\left\|
\epsilon - \epsilon_\theta(z_t, t \,;\, \hat{\mathbf{q}}, c)
\right\|_2^2
\right].
\label{eq:supp_diff}
\end{equation}
Here $\hat{\mathbf{q}}$ denotes the diffusion-conditioning query stream (after cross-attention refinement) and $c=\psi(h_c)$ is the projected expert-token representation used for timestep modulation.

\subsection{Trainable Parameters and Freezing Strategy}
\label{sec:supp_freeze}

\paragraph{Stage I (perception).}
We freeze the backbone and update only: C-MoE branch parameters, the expert token $\tau_c$, and its lightweight prediction head.

\paragraph{Stage II (generation).}
We freeze the C-MoE parameters learned in Stage I, and train: diffusion-side parameters, learnable queries, and $\tau_c$ (plus its head) to accommodate the domain shift of pixelated grayscale references.

\paragraph{Hyperparameters and hardware setup.}
We summarize the hardware setup and key training hyperparameters in Table~\ref{tab:supp_hparams}. We adopt a step-wise linear warmup followed by cosine decay, with weight decay $0.05$.

\subsection{Computational Overhead and Inference Latency}
\label{sec:supp_latency}
Our Composition-MoE (C-MoE) naturally leverages sparse routing to boost model capacity without heavily taxing active computation. Specifically, introducing the diverse compositional experts adds $\sim$8B parameters (a 22\% static VRAM footprint increase). However, during inference, the router activates only a small subset of experts per token. As a result, the \emph{active} parameters and FLOPs per forward pass barely increase. Moreover, since these activated experts execute in parallel instead of deepening the network architecture, the inference latency remains virtually identical to the base unified backbone (e.g., $3.25$s vs. $3.38$s per image on a single NVIDIA A100 GPU). This design elegantly scales learning capacity while maintaining rigorous inference efficiency.

% -------------------------------------------------
\section{Evaluation Metrics}
\label{sec:supp_metrics}

\subsection{Average Precision (AP) and mAP}
\label{sec:supp_ap}

For each category $k$, we compute AP from the precision-recall curve induced by ranking samples using the predicted score $\hat{s}^{(k)}$.
We report mAP as the mean over all categories:
\begin{equation}
\mathrm{mAP}=\frac{1}{C}\sum_{k=1}^{C}\mathrm{AP}_k.
\end{equation}

\subsection{FID and CLIPScore}
\label{sec:supp_fid_clip}

We report FID between generated images and the evaluation set following \cite{heusel2017gans}.
CLIPScore is computed following \cite{hessel-etal-2021-clipscore} by measuring image-text similarity under a pretrained CLIP model.

\subsection{Composition Consistency and Expert-Token Similarity}
\label{sec:supp_comp_metrics}

\paragraph{Composition consistency (Comp-Cons.).}
Let $\hat{y}^{\mathrm{ref}}, \hat{y}^{\mathrm{gen}}\in\{0,1\}^{C}$ be the predicted multi-label vectors for the reference and generated images (predicted by COMPASS).
We measure composition transfer consistency via a set-level agreement:
\begin{equation}
\mathrm{CompCons}
=
\frac{1}{|\mathcal{D}|}\sum_{(I^{\mathrm{ref}}, I^{\mathrm{gen}})\in\mathcal{D}}
\frac{\left|\hat{y}^{\mathrm{ref}}\wedge \hat{y}^{\mathrm{gen}}\right|}{\left|\hat{y}^{\mathrm{ref}}\vee \hat{y}^{\mathrm{gen}}\right|+\epsilon},
\label{eq:supp_compcons}
\end{equation}
where $\wedge/\vee$ denote element-wise AND/OR and $\epsilon$ avoids division by zero. 

\paragraph{Expert-token similarity.}
For an image $I$ and a fixed composition-prediction prompt template, let
$H(I)\in\mathbb{R}^{N\times d}$ denote the final-layer hidden states of the unified LMM,
and let $p_c(I)$ be the (unique) position index of the expert token $\tau_c$ in the resulting token sequence.
We define the expert-token representation as
$h_c(I)=H(I)_{p_c(I)}\in\mathbb{R}^{d}$.
For a reference image $I^{\mathrm{ref}}$ and a generated image $I^{\mathrm{gen}}$,
we report the cosine similarity between their expert-token representations:
\begin{equation}
\cos\!\left(h_c^{\mathrm{ref}},h_c^{\mathrm{gen}}\right)
=
\frac{\langle h_c(I^{\mathrm{ref}}),\, h_c(I^{\mathrm{gen}})\rangle}
{\|h_c(I^{\mathrm{ref}})\|_2\,\|h_c(I^{\mathrm{gen}})\|_2},
\end{equation}
where $\langle\cdot,\cdot\rangle$ is the standard inner product and $\|\cdot\|_2$ is the Euclidean norm.
If multiple instances of $\tau_c$ occur (e.g., multi-turn prompts), we compute $h_c(I)$ by averaging the
corresponding token states before evaluating the cosine similarity.

\begin{table}[t]
\centering
\caption{Training and hardware configuration. Unless specified, the same setting is used for both Stage~I (perception) and Stage~II (generation).}
\label{tab:supp_hparams}
\setlength{\tabcolsep}{6pt}
\renewcommand{\arraystretch}{1.08}
\begin{tabular}{lc}
\toprule
Item & Value \\
\midrule
Hardware (GPU) & 8$\times$ NVIDIA A100 \\
Batch size per GPU & 16 \\
Global batch size & 128 \\
Optimizer & AdamW \\
LR scheduler & step-wise linear warmup \& cosine decay \\
Init LR & $2\times 10^{-5}$ \\
Min LR & $1\times 10^{-6}$ \\
Warmup LR & $1\times 10^{-6}$ \\
Warmup steps & 100 \\
Weight decay & 0.05 \\
Training length (Stage I) & 10 epochs \\
Training length (Stage II) & 10 epochs \\
Loss weights $(\lambda_{\mathrm{cls}},\,\lambda'_{\mathrm{cls}})$ & (1.0, 1.0) \\
Pixelation (blocks wide) & 32 \\
Pixelation resize kernel & nearest neighbor (down \& up) \\
\bottomrule
\end{tabular}
\end{table}

\paragraph{Independent proxy for unbiased evaluation.}
Defining standard quantitative metrics for composition-guided generation is inherently challenging. While we initially utilized our own perception module due to the lack of comprehensive 11-class taxonomy models, doing so may introduce circular dependency bias. To ensure a rigorous and impartial assessment, we additionally revamp our evaluation protocol using an independent third-party proxy model (\href{https://github.com/zhiyuanyou/PhotoFramer-Assessment}{PhotoFramer} \cite{you2025photoframer}, finetuned on external datasets) to compute the composition consistency metrics (Comp-Cons. and $\cos(h_c^{\mathrm{ref}}, h_c^{\mathrm{gen}})$). 

Furthermore, we emphasize that true disentanglement must be evaluated by jointly analyzing Comp-Cons. and CLIPScore. A trivial model that merely copies the reference image's content would inflate the consistency score but severely degrade the text-aligned CLIPScore. As demonstrated in Table~\ref{tab:gen_compare_updated}, COMPASS uniquely excels in both metrics across datasets. This independent third-party assessment confirms that our model successfully isolates abstract structural intents with minimal semantic leakage, proving the improvements are genuine rather than self-evaluation artifacts.

\begin{table}[h]
\vspace{-2mm}
\centering
\caption{Quantitative results on composition-guided generation evaluated by the independent proxy (PhotoFramer \cite{you2025photoframer}). Reported as in-domain / out-of-distribution (OOD).}
\label{tab:gen_compare_updated}
\resizebox{\columnwidth}{!}{
\begin{tabular}{lcccc}
\toprule
Method & FID $\downarrow$ & CLIPScore $\uparrow$ & Comp-Cons. $\uparrow$ & $\cos(h_c^{\mathrm{ref}}, h_c^{\mathrm{gen}})$ $\uparrow$ \\
\midrule
Step1X-Edit \cite{liu2025step1x-edit} & \textbf{7.5} / \textbf{10.2} & 0.72 / 0.68 & 0.64 / 0.56 & 0.54 / 0.63 \\
BAGEL \cite{deng2025emerging} & 7.8 / 10.9 & 0.76 / 0.70 & 0.59 / 0.51 & 0.51 / 0.60 \\
Ming-Lite-Uni \cite{gong2025minglite} & 9.6 / 12.8 & 0.64 / 0.57 & 0.55 / 0.62 & 0.49 / 0.58 \\
\midrule
COMPASS (ours) & 8.2 / 10.6 & \textbf{0.83} / \textbf{0.75} & \textbf{0.71} / \textbf{0.77} & \textbf{0.60} / \textbf{0.68} \\
\bottomrule
\end{tabular}}
\vspace{-2mm}
\end{table}

% \subsection{Head-only Classification}
% \label{sec:supp_headonly}

% We additionally report head-only classification results using the lightweight classifier attached to $\tau_c$.
% \paragraph{(Option A) Single-label accuracy.}
% For datasets/splits where a single dominant composition label is used, we take $\arg\max$ over logits and report top-1 accuracy.
% \paragraph{(Option B) Multi-label exact match / F1.}
% For multi-label evaluation, we can report exact-match accuracy or micro/macro-F1 after thresholding sigmoid outputs.
% (Choose the one matching your experimental protocol and fill in details.)

% -------------------------------------------------
\section{Additional Experimental Results}
\label{sec:supp_extra_exp}

\subsection{Head-only Classification}
Since $\tau_c$ is explicitly trained with the composition classification loss (Eq.~\eqref{eq:weighted-bce}), its attached lightweight head can be directly used as a standalone composition classifier. We therefore report head-only classification accuracy on KU-PCP \cite{lee2018photographic} in Table~\ref{tab:kupcp_acc}, the most widely used benchmark in prior composition classification works, and compare with KU-PCP~\cite{lee2018photographic} and CACNet~\cite{hong2021composing}. COMPASS achieves the best accuracy, confirming that the expert-token representation provides a strong and transferable encoding of composition intent.

\begin{table}[t]
\centering
\caption{Classification accuracy on the KU-PCP test set.}
\label{tab:kupcp_acc}
\begin{tabular}{lc}
\toprule
Method & Acc. (\%) \\
\midrule
KU-PCP \cite{lee2018photographic} & 87.9 \\
CACNet \cite{hong2021composing} & 88.4 \\
COMPASS (ours) & \textbf{90.6} \\
\bottomrule
\end{tabular}
\end{table}

% \subsection{Ablation Records}
% \label{sec:supp_ablation}

% Table~\ref{tab:supp_ablation} provides the numerical ablation results referenced in the main paper.

% \begin{table}[t]
% \caption{Ablation study on composition-guided generation. $\star$ denotes ``same as the previous row'' with additional removals.
% $^{\mathrm{P}}$ and $^{\mathrm{G}}$ indicate perception-stage and generation-stage ablations, respectively.}
% \label{tab:supp_ablation}
% \centering
% \small
% \setlength{\tabcolsep}{3.5pt}
% \renewcommand{\arraystretch}{1.08}
% \begin{tabularx}{\columnwidth}{@{}>{\raggedright\arraybackslash}Xccccc@{}}
% \toprule
% Variant &
% mAP $\uparrow$ &
% FID $\downarrow$ &
% CLIPScore $\uparrow$ &
% Comp-Cons. $\uparrow$ &
% $\cos(h_c^{\mathrm{ref}}, h_c^{\mathrm{gen}})$ $\uparrow$ \\
% \midrule
% Full (COMPASS) & \textbf{90.6} & 8.2 & \textbf{0.83} & \textbf{0.74} & \textbf{0.62} \\
% w/o C-MoE$^{\mathrm{P}}$ & 78.2 & -- & -- & -- & -- \\
% w/o $\tau_c$$^{\mathrm{P}}$ & 71.1 & -- & -- & -- & -- \\
% w/o cross attention$^{\mathrm{G}}$ & -- & \textbf{8.0} & 0.79 & 0.64 & 0.56 \\
% $\star$ + w/o mask$^{\mathrm{G}}$ & -- & 8.6 & 0.72 & 0.59 & 0.48 \\
% $\star$ + w/o $\tau_c$$^{\mathrm{G}}$ & -- & 8.3 & 0.75 & 0.53 & 0.42 \\
% \bottomrule
% \end{tabularx}
% \end{table}

\subsection{Out-of-Distribution (OOD) Generalization}
\label{sec:supp_ood}

To further validate the generalization capabilities of COMPASS beyond the \textsc{Comp-11} domain, we conduct a zero-shot evaluation on an independent OOD subset from the UNIAA dataset. We construct this subset with 1,424 labeled perception samples (maintaining a strict 1:1 positive-to-negative ratio per category) and 354 distribution-consistent generation pairs. Because UNIAA targets general aesthetics, it lacks the massive scale and fine-grained layout annotations (e.g., ``Fill the Frame'', ``Pattern'') of our work, conversely highlighting the indispensability of the \textsc{Comp-11} dataset.

As shown in Table~\ref{tab:ood_precision}, while the impartial proxy PhotoFramer \cite{you2025photoframer} demonstrates competent zero-shot perception, COMPASS still achieves the highest overall mAP without any finetuning. Furthermore, the superior quantitative generation results (Table~\ref{tab:gen_compare_updated}) and qualitative visualizations (Figure~\ref{fig:disentanglement}) on this OOD set confirm the model's robustness. This strong cross-dataset performance proves that our model learns generalized compositional abstractions rather than overfitting the training distribution.

\begin{table}[h]
\vspace{-2mm}
\centering
\caption{Zero-shot perception evaluation on the OOD UNIAA subset.}
\label{tab:ood_precision}
\resizebox{\columnwidth}{!}{
\begin{tabular}{lcccccccccc}
\toprule
Method & RoT & Center & Hori. & Vert. & Diag. & Curved & Tri. & Sym. & Lead. & \textbf{Total} \\
\midrule
\textit{\#Samples} & \textit{410} & \textit{232} & \textit{64} & \textit{26} & \textit{234} & \textit{86} & \textit{48} & \textit{264} & \textit{60} & \textit{1424} \\
\midrule
AesExpert \cite{huang2024aesexpert} & 50.0 & 53.2 & 50.0 & 48.0 & 50.0 & 50.0 & 50.0 & 50.0 & 50.0 & 50.5 \\
Qwen3-VL \cite{qwen3vl_2025} & \underline{72.6} & 64.4 & 40.6 & 70.0 & 59.0 & 66.2 & 91.3 & 86.4 & 58.0 & 69.7 \\
InternVL3 \cite{zhu2025internvl3} & 54.6 & 68.8 & 33.3 & 41.7 & 71.1 & 67.2 & 82.1 & \underline{89.4} & 57.7 & 66.7 \\
Janus-Pro \cite{wu2025janus} & 50.2 & 55.7 & \underline{50.9} & 52.2 & 62.2 & 55.8 & 61.8 & 71.3 & 50.0 & 57.8 \\
BAGEL \cite{deng2025emerging} & 53.7 & 66.0 & 38.1 & 50.0 & 64.8 & 68.9 & 81.5 & 80.9 & 74.4 & 64.5 \\
Ming-Lite-Uni \cite{gong2025minglite} & 58.6 & 71.5 & 49.1 & 40.0 & 69.0 & 70.0 & 91.3 & 82.4 & 66.7 & 68.2 \\
PhotoFramer \cite{you2025photoframer} & 69.5 & \underline{77.3} & 48.3 & \underline{80.0} & \underline{80.1} & \underline{80.7} & \underline{92.1} & 83.5 & \underline{79.6} & \underline{76.2} \\
\midrule
COMPASS (ours) & \textbf{92.4} & \textbf{82.2} & \textbf{50.9} & \textbf{100.0} & \textbf{91.7} & \textbf{97.1} & \textbf{100.0} & \textbf{91.0} & \textbf{92.0} & \textbf{89.2} \\
\bottomrule
\end{tabular}}
\vspace{-2mm}
\end{table}

\subsection{Additional Qualitative Results and Failure Cases}
\label{sec:supp_qual}

In this section, we provide additional qualitative comparisons for both composition perception and composition-guided generation.

\begin{figure*}[t]
  \centering
  \includegraphics[width=\textwidth]{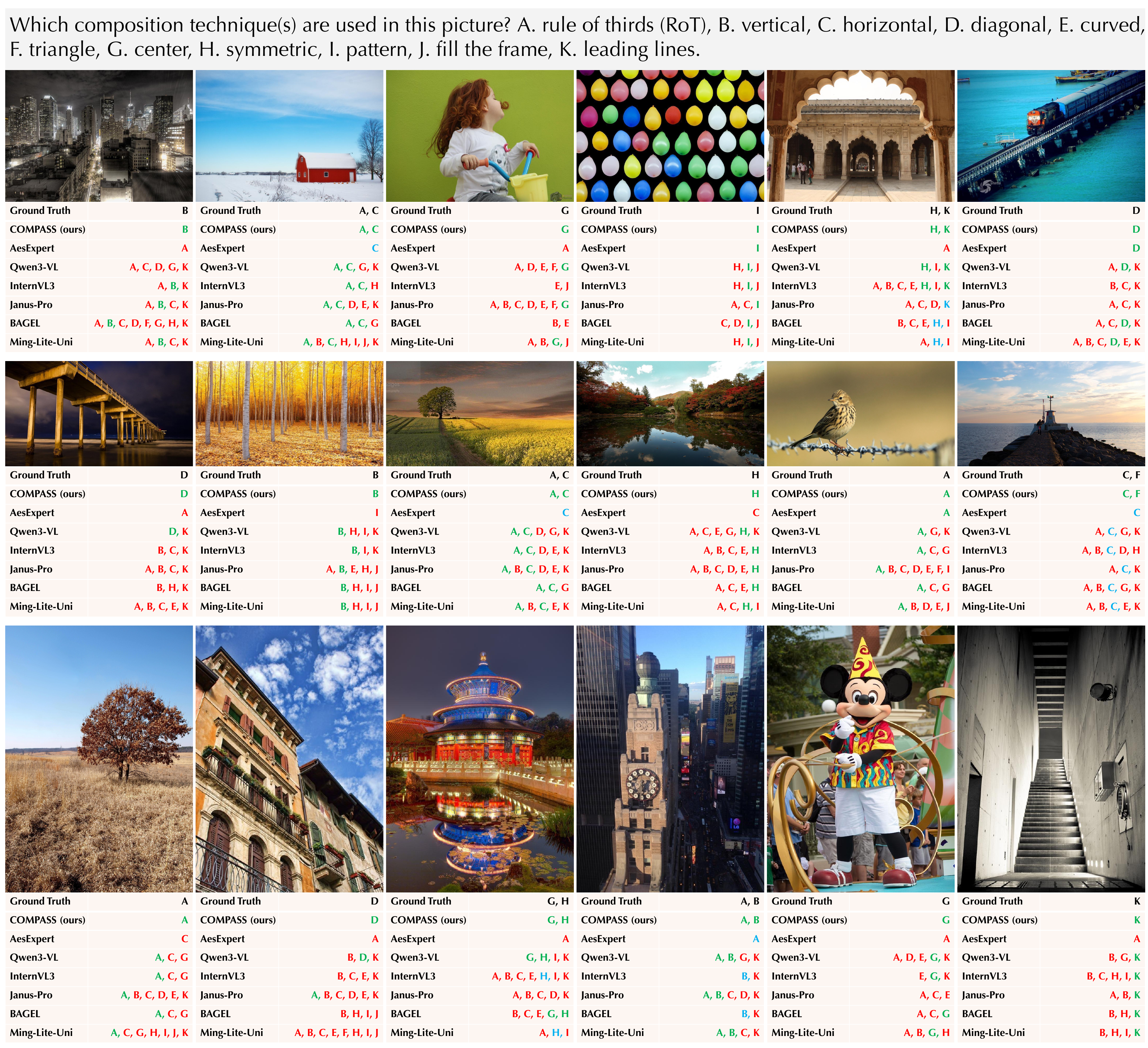}
  \caption{Qualitative comparison on multiple-choice composition identification.
  We color-code each option for intuitive reading: green = correctly selected, red = incorrectly selected, blue = missed
  ground-truth option. Compared with strong baselines, COMPASS selects composition techniques more accurately and avoids
  ``hedging'' by over-selecting frequent labels.}
  \label{fig:supp_mcq}
\end{figure*}

\begin{figure*}[htbp]
  \centering
  \includegraphics[width=\textwidth]{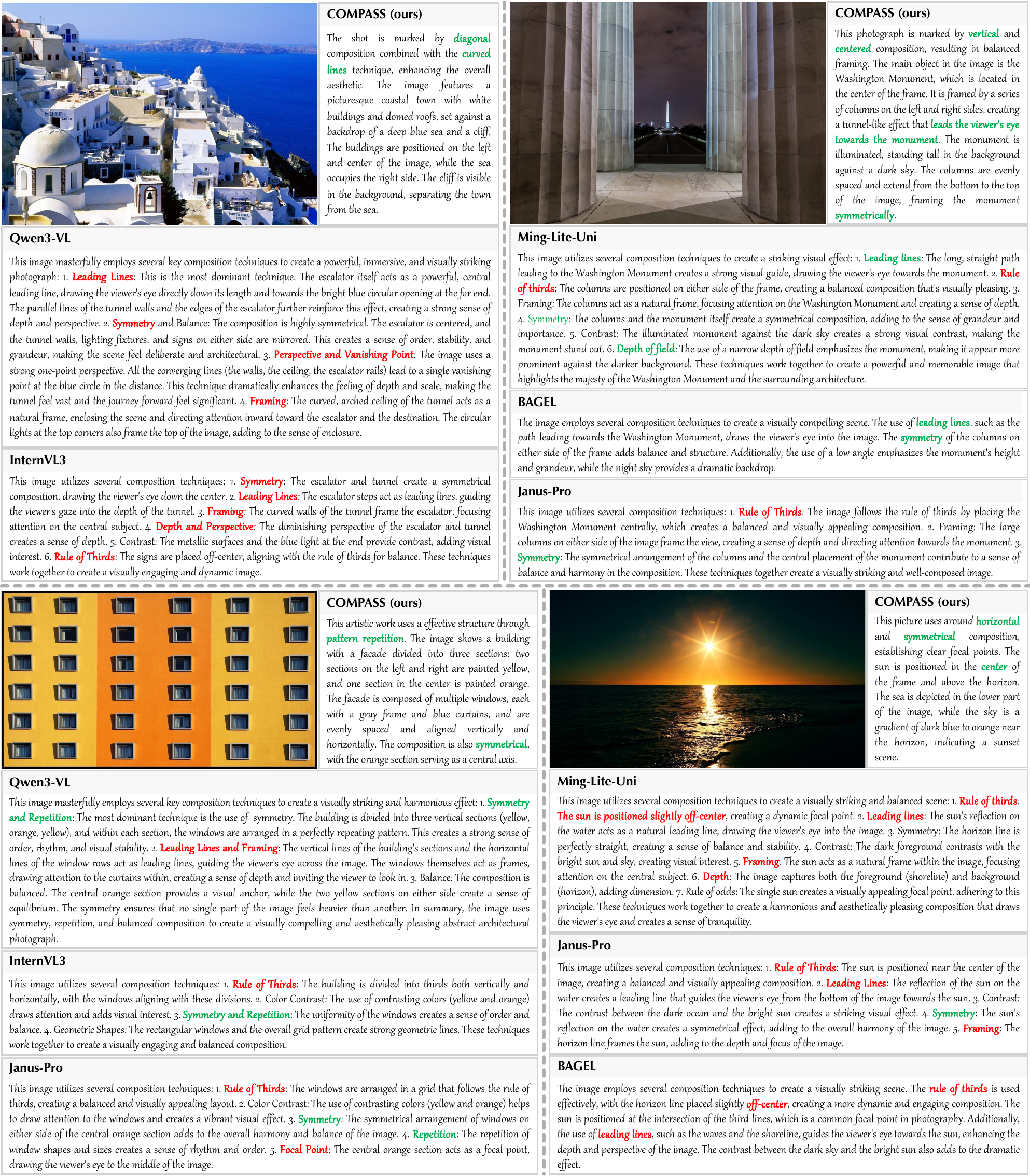}
  \caption{Qualitative comparison on VQA-style compositional analysis.
  Highlighted technique words indicate whether the mentioned composition cues match the ground truth (green) or reflect hallucinated/unsupported claims (red). COMPASS produces more grounded and composition-consistent analyses, whereas other LMM/unified baselines often repeat high-frequency concepts independent of the actual layout.
  }
  \label{fig:supp_vqa}
\end{figure*}

\paragraph{Composition perception results.}
We provide additional qualitative comparisons on two perception formats included in \textsc{Comp-11}: multiple-choice
composition identification and free-form compositional analysis (VQA). As shown in
Figure~\ref{fig:supp_mcq} and \ref{fig:supp_vqa}, COMPASS more consistently grounds its predictions in the visual evidence and
selects/explains the correct composition techniques, while strong LMM/unified baselines frequently fall back to
generic ``composition buzzwords'' (e.g., repeatedly mentioning RoT, leading lines, or framing-related cues) regardless of
whether such patterns are present.

\emph{Multiple-choice comparisons.}
Figure~\ref{fig:supp_mcq} shows representative multiple-choice discrimination cases.
A common failure mode of existing models is over-selection: they tend to tick several popular options as a hedge,
which often looks like a lucky guess rather than a grounded decision.
For example, RoT (A) and Center (G) are frequently selected together even when the image primarily reflects one dominant subject
placement, suggesting limited discrimination over fine-grained layout intent.
In contrast, COMPASS typically selects a compact set of options that aligns with the ground-truth labels, indicating
stronger composition-sensitive perception.

\emph{Free-form VQA comparisons.}
In this setting, models are asked to ``analyze the composition of the image''. Figure~\ref{fig:supp_vqa} compares open-ended compositional analyses.
Most baselines exhibit a ``list-and-describe'' behavior: they enumerate common techniques with plausible-sounding
justifications, but the mentioned cues are often weakly supported by the image or even mutually inconsistent within the
same response.
COMPASS, by contrast, identifies the correct techniques with more image-grounded explanations (e.g., explicitly linking
dominant geometric cues and spatial organization to the predicted composition intent), demonstrating that the learned
composition expertise is not only classifiable but also explainable.

\paragraph{Composition-guided generation results.}
We provide additional qualitative results to complement the quantitative evaluation in the main paper. Our goal is to diagnose whether a model can effectively extract composition intent from a reference image and realize this intent while synthesizing new semantics from text, rather than copying residual reference appearance. To this end, we present three sets of analyses: (i) an evaluation of composition-content disentanglement on out-of-distribution (OOD) domains, (ii) extended qualitative comparisons against baseline models, and (iii) an ablation demonstrating that input-side pixelization alone is insufficient without model-side intent grounding.

\begin{figure}[h]
\centering
\includegraphics[width=\linewidth]{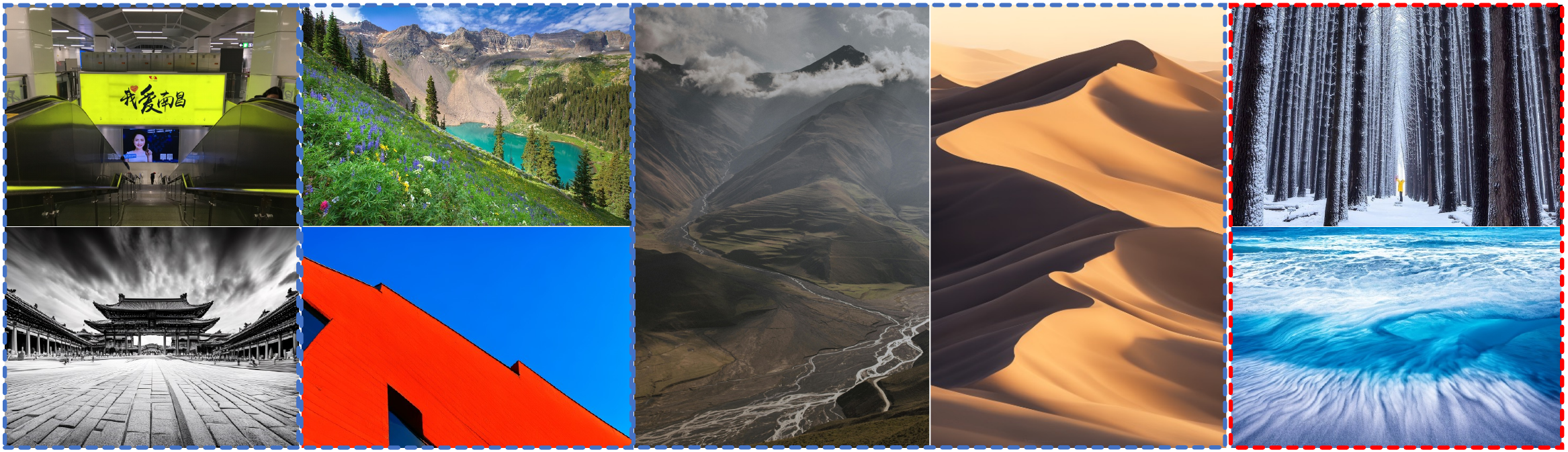}
\caption{Qualitative evaluation of disentanglement on OOD data.}
\label{fig:disentanglement}
\end{figure}

\emph{Disentanglement on OOD domains.} 
Beyond in-domain evaluations, Figure~\ref{fig:disentanglement} (blue boxes) demonstrates COMPASS's composition-content disentanglement on OOD data. COMPASS robustly extracts layouts across domains, from non-aesthetic everyday photos to non-photographic watercolors. To prove it isolates pure layout from strongly coupled content, consider the diagonal example: a mountain slope reference versus a ``red building wall'' target. Neither inherently dictates a diagonal composition. Yet, COMPASS precisely extracts \textit{only} the geometric intent without leaking ``mountain'' semantics into the target architecture domain.

\begin{figure*}[t]
  \centering
  \includegraphics[width=\textwidth]{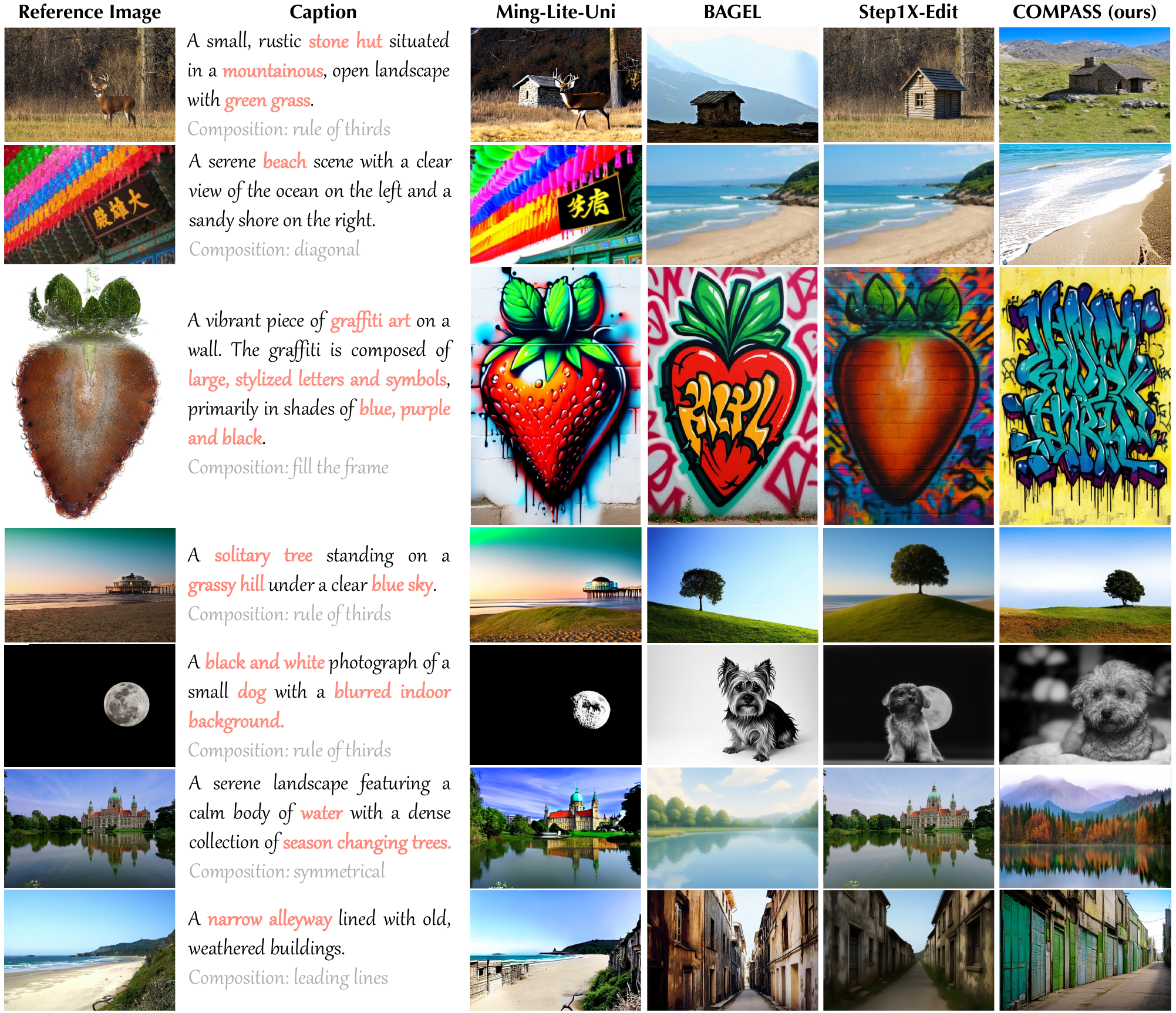}
  \caption{Additional qualitative comparison on composition-guided generation. Gray annotations under captions are for intuitive illustration only and are not used as model inputs.}
  \label{fig:supp_gen_more}
\end{figure*}

\emph{Additional comparisons against baselines.}
Figure~\ref{fig:supp_gen_more} extends the qualitative comparisons in the main paper.
The results show that COMPASS exhibits a capability largely missing from current image editing and unified models: it can infer the compositional intent from the reference and integrate this intent into the prompt-specified content, producing generations that are more composition-consistent and semantically coherent.
In contrast, baselines often either drift toward prompt content with weak composition control or exhibit unintended reference leakage and over-dependence on the reference appearance.

\begin{figure*}[t]
  \centering
  \includegraphics[width=\textwidth]{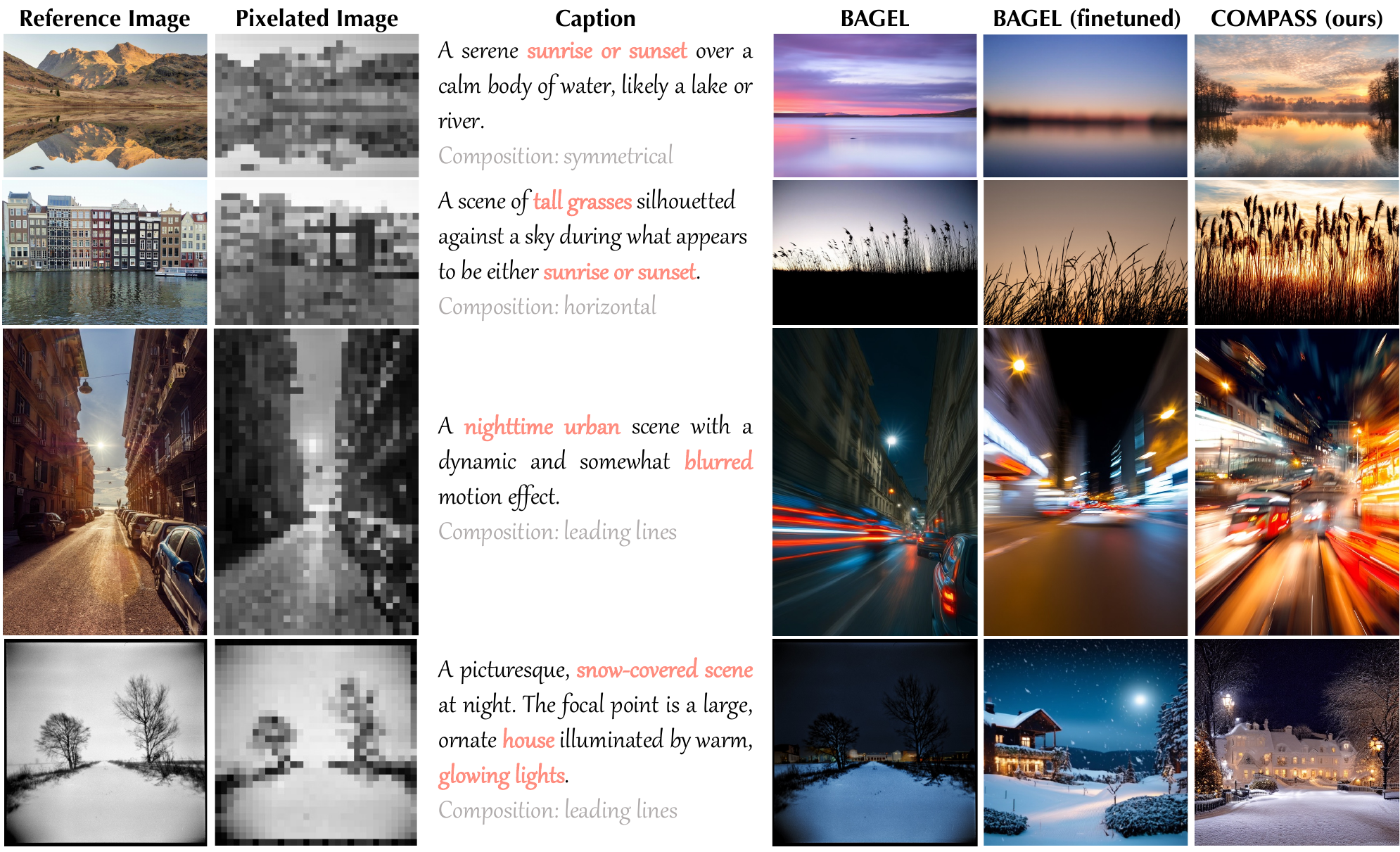}
  \caption{Ablation on structural bottlenecking via pixelization. Finetuning a unified model with pixelized references alone does not reliably learn composition-intent transfer, highlighting the necessity of expert-token-based intent grounding and generation-time conditioning.}
  \label{fig:supp_bagel_pix}
\end{figure*}

\emph{Why pixelation alone is insufficient.}
To isolate the effect of input-side structural bottlenecking, Figure~\ref{fig:supp_bagel_pix} compares (i) the original BAGEL using raw reference images, (ii) BAGEL finetuned with pixelized references on \textsc{Comp-11}, and (iii) COMPASS.
While pixelization weakens the reference semantic pathway, we observe that pixelization-only finetuning does not reliably yield composition transfer.
This supports our central motivation: after removing semantic appearance cues, the model still needs an explicit composition inference and summarization process (anchored by the expert token) and a generation-time mechanism to inject the inferred intent as a global control signal.
In other words, suppressing reference semantics is necessary but not sufficient; composition-intent controllability requires model-side intent grounding and conditioning.

% \paragraph{Failure cases.}
% Figure~\ref{fig:supp_failure} summarizes representative failure cases of COMPASS.
% We observe two recurring patterns.
% First, the model may preserve the composition category but mismatch orientation (e.g., left/right subject placement for RoT, or opposite diagonal direction), suggesting that the dominant signal is still category-level intent while the remaining coarse layout cues in the pixelized reference may be insufficient to fully determine directional instantiation.
% Second, the model may misinterpret composition when the image exhibits strong perspective/blur cues: e.g., a scene that resembles a diagonal layout due to background leading structure, while the true intent is RoT with a focused subject (the left bottom example).
% These cases indicate that improving depth-of-field and 3D spatial reasoning for composition remains an important direction.

\begin{figure}[t]
  \centering
  \includegraphics[width=\columnwidth]{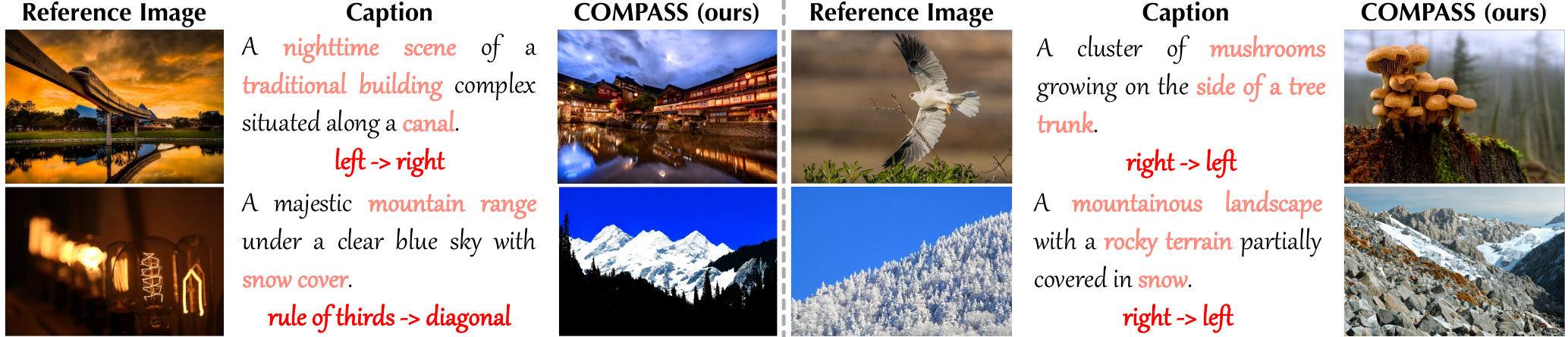}
  \caption{Failure cases of COMPASS. Common issues include orientation mismatch under the correct composition type, and occasional misclassification when perspective/defocus cues dominate perceived structure.}
  \label{fig:supp_failure}
\end{figure}

\paragraph{Failure cases and semantic priors.}
We observe two main categories of failure patterns, as summarized in Figure~\ref{fig:supp_failure} and the red box of Figure~\ref{fig:disentanglement}.

First, regarding spatial and perception errors (Figure~\ref{fig:supp_failure}), the model may preserve the correct composition category but mismatch the orientation (e.g., left/right subject placement for RoT), or occasionally misclassify the layout when the image exhibits strong perspective or defocus cues.

Second, regarding semantic-layout conflicts (red box in Figure~\ref{fig:disentanglement}), when the target layout fundamentally contradicts the natural semantics (e.g., forcing a ``vertical'' layout onto ``ocean waves'' that are inherently horizontal), COMPASS typically prioritizes physical realism over strict structural adherence. While technically a failure in conditional control, this trade-off gracefully prevents corrupted artifacts (e.g., rendering oceans as vertical waterfalls), demonstrating that the model avoids blindly forcing layouts against natural logic. Handling these extreme semantic-layout conflicts remains an interesting direction for future work.

% -------------------------------------------------

\section{Broader Impact of the Staged Pipeline}
\label{sec:supp_generalizability}

While instantiated specifically on ``composition'' in this paper, our work fundamentally tackles a broader challenge in visual generation: injecting abstract, high-level visual intents into unified Large Multimodal Models (LMMs). To this end, our staged pipeline provides a foundational and reusable recipe. The stage-wise design---perception warm-up followed by generation grounding---introduces continuous control signals without causing catastrophic forgetting of general capabilities. 

Crucially, this identical pipeline can be seamlessly adapted to ground other subjective and abstract concepts (e.g., scene lighting, emotional tones, cinematic camera movements) simply by swapping the expert data branch. Therefore, beyond offering a mere downstream application, we contribute a generalizable framework for abstract visual control. To empower future community research and facilitate continuous improvements on this foundational paradigm, we open-source our complete codebase, trained models, and custom metric scripts.

\section{Ethics Considerations}
\label{sec:ethics}

\paragraph{Dataset sourcing and licensing.}
\textsc{Comp-11} is constructed by aggregating images from public aesthetic benchmarks (AVA \cite{murray2012ava}, TAD66K \cite{he2022rethinking}, CADB \cite{zhang2021cadb}, KU-PCP \cite{lee2018photographic}).
We will comply with the original datasets' licenses and terms of use, and we will provide clear attribution and usage conditions in the release package.
If any source imposes restrictions that prevent redistribution, we will release only the derived annotations/metadata and provide scripts to reproduce the dataset from the original sources when feasible.

\paragraph{Privacy and sensitive content.}
The underlying benchmarks may contain images of people or potentially sensitive scenes. We emphasize that our images strictly originate from established public benchmarks with built-in sensitive content screening, rather than raw or uncurated web scraping. While we do not collect private data beyond what is already public, we acknowledge the inherent risk of unintentionally propagating personally identifiable information (PII). To mitigate this, our release package includes comprehensive guidance on responsible use, and we implement a strict takedown protocol for any problematic content upon request.

\paragraph{Annotation labor and annotator welfare.}
Expert labeling involves subjective judgments and may incur cognitive load.
We follow a multi-annotator protocol with consensus resolution to reduce individual pressure and bias.
If we release additional annotation guidelines and tooling, we will document the process and recommend fair compensation and reasonable workload management for any future extensions.

\paragraph{Potential misuse.}
Composition-aware generation can be used to create persuasive or misleading imagery.
Our work focuses on compositional intent (layout control) rather than identity manipulation or targeted deception, but the broader capability could be misused.
We will include a responsible-use statement and encourage downstream users to follow applicable platform policies and local regulations for synthetic media disclosure.

\paragraph{Bias and representativeness.}
Aesthetic and composition preferences can be culturally dependent.
Because the source benchmarks may over-represent certain photographic styles, our taxonomy and annotations might not fully cover diverse cultural composition norms.
We will document dataset composition and known limitations, and we encourage future work to expand coverage across cultures, genres, and contexts.

\paragraph{Environmental impact and reproducibility.}
Training unified multimodal models and diffusion generators can be compute-intensive.
To reduce unnecessary computation compared to full-model fine-tuning, our design emphasizes parameter-efficient expertization (C-MoE branch and lightweight modules) and stage-wise training.
We will report key training configurations in the supplementary material for transparency and reproducibility, and plan to release the dataset and code once organized.

\end{document}